\documentclass[sigconf,nonacm,10pt]{acmart}

\makeatletter
\def\@ACM@checkaffil{
    \if@ACM@instpresent\else
    \ClassWarningNoLine{\@classname}{No institution present for an affiliation}%
    \fi
    \if@ACM@citypresent\else
    \ClassWarningNoLine{\@classname}{No city present for an affiliation}%
    \fi
    \if@ACM@countrypresent\else
        \ClassWarningNoLine{\@classname}{No country present for an affiliation}%
    \fi
}
\makeatother

\usepackage{graphicx}
\usepackage{amsmath}
\usepackage{bm}
\usepackage{xcolor}
\usepackage{colortbl}
\usepackage{soul}
\usepackage[linesnumbered,ruled,vlined]{algorithm2e}
\usepackage[export]{adjustbox}
\usepackage{rotating}
\usepackage{array}
\usepackage{tabularx}
\usepackage{tikz}
\usepackage{float}
\usepackage[noabbrev, capitalise]{cleveref}
\usepackage{multirow}
\usepackage{wrapfig}
\usepackage[font=footnotesize,labelfont=bf]{caption, subcaption}
\usepackage{caption}
\usepackage{listings}
\newcommand*\darkcircled[1]{\tikz[baseline=(char.base)]{\node[shape=circle,draw,inner sep=1.2pt, white, fill=black] (char) {#1};}}

\definecolor{brightmaroon}{rgb}{0.76, 0.13, 0.28}
\setul{1pt}{0.5pt} 

\begin{document}


\title {\large {A Smart-Glasses for Emergency Medical Services via Multimodal Multitask Learning}
}




\author{\small Liuyi Jin$^1$, Pasan Gunawardena$^1$*, Amran Haroon$^1$*, Runzhi Wang$^2$, Sangwoo Lee$^1$, Radu Stoleru$^1$, \\ Michael Middleton$^3$, Zepeng Huo$^4$, Jeeeun Kim$^1$, Jason Moats$^5$}

\affiliation{\footnotesize $^1$Computer Science and Engineering,  $^2$Electrical and Computer Engineering, Texas A\&M University \\ $^3$Texas A\&M University Emergency Medical Services (EMS), $^4$Biomedical Data Science, Stanford University, $^5$Texas A\&M School of Public Health }

\email{{liuyi, pgunawardena, amran.haroon, runzhi354, swlee, stoleru, mmiddleton, jeeeun.kim, jbmoats}@tamu.edu, frazierhuo@gmail.com}

\thanks{*equal secondary contribution}

%


\begin{abstract}

Emergency Medical Technicians (EMTs) operate in high-pressure environments, making rapid, life-critical decisions under heavy cognitive and operational loads. We present \textbf{EMSGlass}, a smart-glasses system powered by \textbf{EMSNet}, the first multimodal multitask model for Emergency Medical Services (EMS), and \textbf{EMSServe}, a low-latency multimodal serving framework tailored to EMS scenarios. EMSNet integrates text, vital signs, and scene images to construct a unified real-time understanding of EMS incidents. Trained on real-world multimodal EMS datasets, EMSNet simultaneously supports up to five critical EMS tasks with superior accuracy compared to state-of-the-art unimodal baselines. Built on top of PyTorch, EMSServe introduces a modality-aware model splitter and a feature caching mechanism, achieving adaptive and efficient inference across heterogeneous hardware while addressing the challenge of asynchronous modality arrival in the field. By optimizing multimodal inference execution in EMS scenarios, EMSServe achieves \textbf{1.9× – 11.7×} speedup over direct PyTorch multimodal inference. 
A user study evaluation with six professional EMTs demonstrates that EMSGlass enhances real-time situational awareness, decision-making speed, and operational efficiency through intuitive on-glass interaction. In addition, qualitative insights from the user study provide actionable directions for extending EMSGlass toward next-generation AI-enabled EMS systems, bridging multimodal intelligence with real-world emergency response workflows.

\end{abstract}

\maketitle
\pagestyle{plain}

\balance

\section{Introduction}
\label{sec:intro}

Emergency Medical Services (EMS) coordinate medical resources, to provide immediate pre-hospital care during emergencies, including large-scale natural disasters (e.g., the California wildfires in January 2025~\cite{CMSAnnounceAssist2025,AssistanceForFires2025}) and family emergency care in rural areas~\cite{RuralEMS2016}. Effective EMS is crucial for saving lives in these critical situations. Emergency Medical Technicians (EMTs), trained in EMS protocols and equipped with essential medical resources, play a central role in delivering such emergency care. However, the EMT profession faces a severe workforce shortage due to high stress levels and long working hours, leading to widespread burnout and attrition~\cite{EMSBurnout2023,EMTStress2022}. These challenges, combined with the complexities of real-time decision-making, make it increasingly difficult for EMTs to ~\ul{\textbf{deliver optimal and timely EMS}}.

\begin{figure}[h]
    \includegraphics[width=\columnwidth]{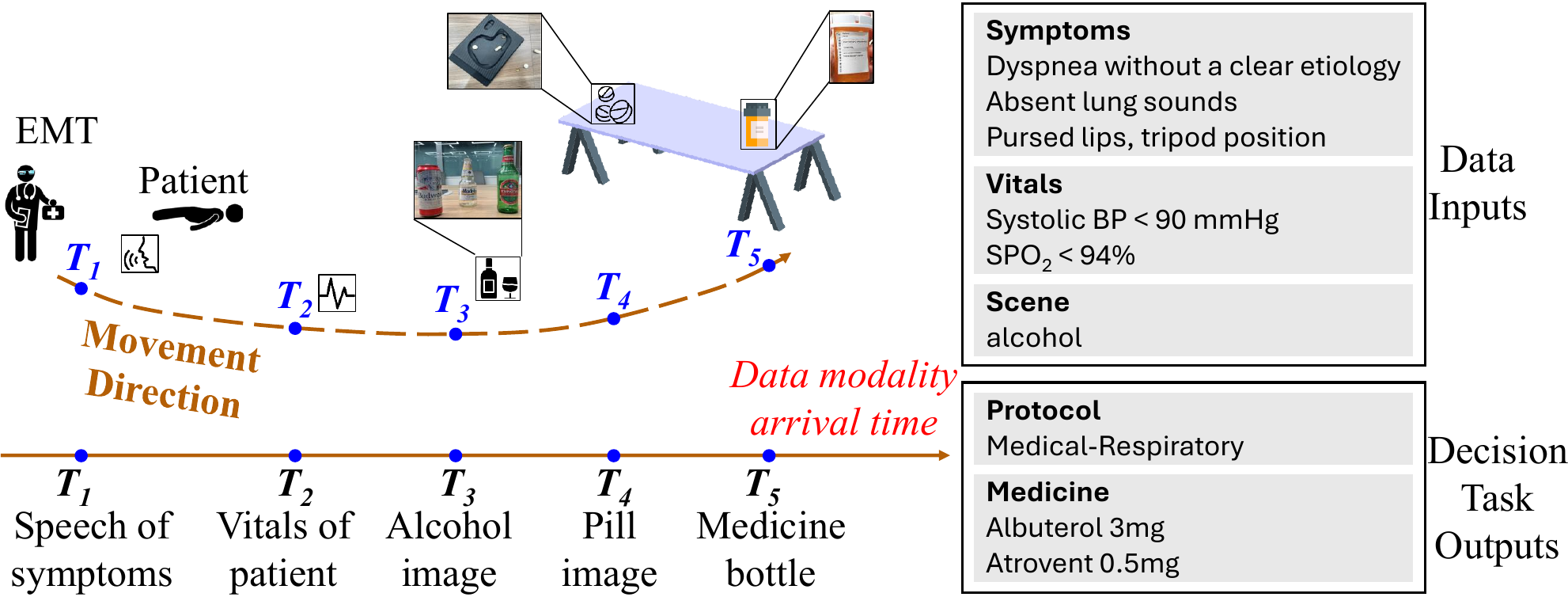}
    \caption{While moving, the EMT perceives multimodal EMS data at different times and selects the protocol ``Medical-Respiratory''.}
    \Description{EMT receives different types of EMS data at different times while moving and selects a medical protocol.}
    \label{fig:different_time_arrival}
\end{figure}

Figure~\ref{fig:different_time_arrival} illustrates the workflow of an EMT in a typical EMS scenario. The EMT first assesses the patient's symptoms, then uses medical equipment to obtain vital signs, and finally evaluates the surrounding environment. These multimodal data--symptoms, vitals, and images in the scene contexts including alcohol and pills--are asynchronously collected and processed in sequence to inform real-time clinical decision-making. As these multimodal data arrived (i.e., perceived by EMTs) at different timestamps, one of the critical tasks for EMTs is to select the appropriate EMS protocol from over 100 available options, which requires matching the asynchronously observed multimodal data with that defined in each protocol. For example, in Figure~\ref{fig:different_time_arrival}, after analyzing the asynchronously perceived multimodal data, the EMT selects the protocol ``Medical-Respiratory,'' which prescribes general clinical interventions including the administration of Atrovent medicine at a quantity of 0.5mg. Making such high-stakes clinical decisions under time constraints is critical for optimal EMS delivery but remains a significant challenge for EMTs.

Mobile assistant systems for EMS have been evolving to harness both cloud and edge (e.g., smart glasses) resources for real-time cognitive assistance and protocol recommendations~\cite{ShuBehaviorIROS2019,PreumCognitiveEMSSigbed2019,PreumTowardsICCPS2018}. For instance, EMSAssist~\cite{EMSAssist2023MobiSys,EMSAssistDemo2023} and CognitiveEMS~\cite{CognitiveEMS2024IoTDI} allow EMTs to wear smart glasses for capturing the voice of patient symptoms. They utilize domain-specific speech-to-text models (e.g., Whisper-tiny) to transcribe the speech and data-driven models (e.g., MobileBERT) to help select protocols. These models are trained in the cloud and deployed on hands-free smart glasses (e.g., Google Glass, Vuzix M4000) and edge servers to address challenges such as unreliable cloud connectivity in disaster scenarios while also enhancing user privacy. Despite these advancements, several clinical and technical challenges remain unresolved:

\textbf{Challenge \#1: Lacking a multimodal model for multiple simultaneous EMS tasks.} On the input side, state-of-the-art (SOTA) systems~\cite{CognitiveEMS2024IoTDI,EMSAssist2023MobiSys} rely solely on patient symptoms but ignore vitals and scene images (Figure~\ref{fig:different_time_arrival}), which are all necessary modalities for comprehensive situational awareness and contextual understanding. This limitation leads to suboptimal EMS delivery by failing to account for multimodal data. On the output side, while EMS protocols dictate general clinical guidance, EMTs must handle diverse tasks including real-time treatment personalization, e.g., extra medication or additional dosage based on patients' rapidly evolving conditions. Restricting focus on the protocol selection in existing systems reduces their practicality and usability, highlighting the need for a multimodal model that supports multiple simultaneous EMS tasks.

\textbf{Challenge \#2: High-latency serving frameworks for multimodal models in EMS scenarios.} Existing multimodal model serving frameworks, such as PyTorch~\cite{pytorch2019} and TensorFlow~\cite{tensorflow2016} used in SOTA assistant systems~\cite{CognitiveEMS2024IoTDI,EMSAssist2023MobiSys}, assume simultaneous availability of all data modalities. However, in the EMS scenario shown in Figure~\ref{fig:different_time_arrival}, smart glasses receive different data modes asynchronously (i.e., asynchronous arrival times) as EMTs move through the scene. Without caching intermediate processing results for early arrived modes of data, directly using multimodal model serving frameworks has to repeatedly process early arrived voice data when later-arriving data like vitals and images become available, leading to redundant computations and higher latency. To achieve timely EMS delivery given the asynchronous arrival of multimodal data, a low-latency multimodal serving framework in EMS scenarios is essential.

\textbf{Challenge \#3: Lacking effective user interaction designs and real-world user studies.}
As illustrated in Figure~\ref{fig:different_time_arrival}, while EMTs move dynamically through emergency scenes, smart glasses continuously receive asynchronous data streams. For each newly arrived data, EMTs expect prompt and reliable recommendation updates to support time-critical decisions. However, the user interface designs in existing EMS assistant systems~\cite{CognitiveEMS2024IoTDI,EMSAssist2023MobiSys} fail to account for either EMT mobility or the variability of data arrival times in practice. Although the behavior tree assistant ~\cite{ShuBehaviorIROS2019,PreumCognitiveEMSSigbed2019,PreumTowardsICCPS2018} supports continuous recommendation updates, its laptop-based interface restricts hands-free operation and ignores the stringent resource constraints of smart glasses. Furthermore, prior works lack real-world user studies with EMTs' end-to-end usage, leaving their effectiveness and usability of interaction designs in real-world EMS scenarios largely unexamined.

In this paper we present EMSGlass, the first multimodal multitask model-enabled smart glasses system for EMS, with the following threefold contributions:

\textbf{(1)} We build \textbf{EMSNet}, the first multimodal multitask model trained on massive, real-world multimodal EMS datasets to simultaneously accomplish five critical EMS tasks: protocol selection, recommendation for medicine type, quantity, dosage, and disease history inference.

\textbf{(2)} We develop \textbf{EMSServe}, the first serving framework for multimodal models to address the challenges introduced by different data arrival times in EMS scenarios. With the adaptive edge-assisted offloading, EMSServe harnesses the computing resources on both the glass and edge servers. Comprehensive evaluations show EMSServe generally outperforms direct usage of PyTorch by 1.9× – 11.7× speedup across diverse edge devices (Google Glass) and edge servers.

\textbf{(3)} We implement a \textbf{seamless on-display user interface and comprehensive user study}. The user interface enables continuous and intuitive recommendation updates on the Google Glass display with lower latencies. The quantitative user study results validate the effectiveness of interaction design. The qualitative results inform actionable system improvements to promote real-world user adoptions.


This paper's data, code, and model will be open-sourced upon acceptance.

\section{Background and Motivation}
\label{sec:background}

\subsection{Multimodal multitask model}
Multimodal multitask models~\cite{visionFoundation2024,DISTMM2024NSDI} are designed to understand and fuse complementary information from multiple sources. These models rely on one or more deep learning submodules to process and integrate data effectively. They aim to generate intermediate unified numeric representations from multimodal inputs for multiple simultaneous tasks across diverse domains, e.g., general healthcare~\cite{huang2020fusion,Lian2024npj,li2019early,kharazmi2018feature}. However, few has explored the feasibility and advantage of such multimodal multitask capability in the EMS domain.

\subsection{Virtual EMS assistants on smart glasses}
Since the release of Google Glass Explorer Edition as a ubiquitous computing platform~\cite{GoogleGlassExplorer2012}, smart glasses have been integrated into various medical applications, including surgical assistance~\cite{pedisurgery2014}, eating behavior monitoring~\cite{eating2015PervasiveHealth}, physiological vitals sensing~\cite{SpiderNie2020,SPIDERSNie2021}, and cognitive state and ocular health tracking~\cite{BlinkMental2025}. Despite their promising hands-free interaction and heads-up display capabilities, the potential of on-glass assistants to enhance EMS delivery remains largely underexplored. Recent systems such as EMSAssist~\cite{EMSAssistDemo2023} and CognitiveEMS~\cite{CognitiveEMS2024IoTDI} have made initial attempts to assist EMTs with protocol selection on smart glasses. However, they either lack multimodal multitasking capabilities or comprehensive real-world user studies, leaving a critical gap between technological advancement and practical adoption in EMS.

\subsection{Object detection and OCR in EMS}
\label{sec:object_detection_ocr_background}

The national EMS dataset (NEMSIS)~\cite{nemsis2015} mandates reporting pill and alcohol presence at EMS scenes due to their strong association with overdose and respiratory emergencies. Accurate detection of these objects can narrow protocol options (e.g., from over 100 to fewer than 5), yet EMS-specific object detectors remain underdeveloped. In parallel, accurate medication administration requires precise Med-Math (e.g., administering $21mg$ of Adrenaline from a $4.2mg/ml$ solution: \( \frac{21mg}{4.2mg/ml} = 5ml \)), but complex unit conversions and vendor-specific concentration increase cognitive load and delay critical care. Smart-glass cameras can automate medication recognition through OCR and barcode scanning, enabling instant extraction of drug names and concentrations, linking to patient history, and reducing errors. Despite its clinical importance, no prior work has focused on developing such integrated object detection and OCR pipelines for EMS.

\section{EMSNet Design}
\label{sec:system_design}

\begin{figure}
    \includegraphics[width=\linewidth]{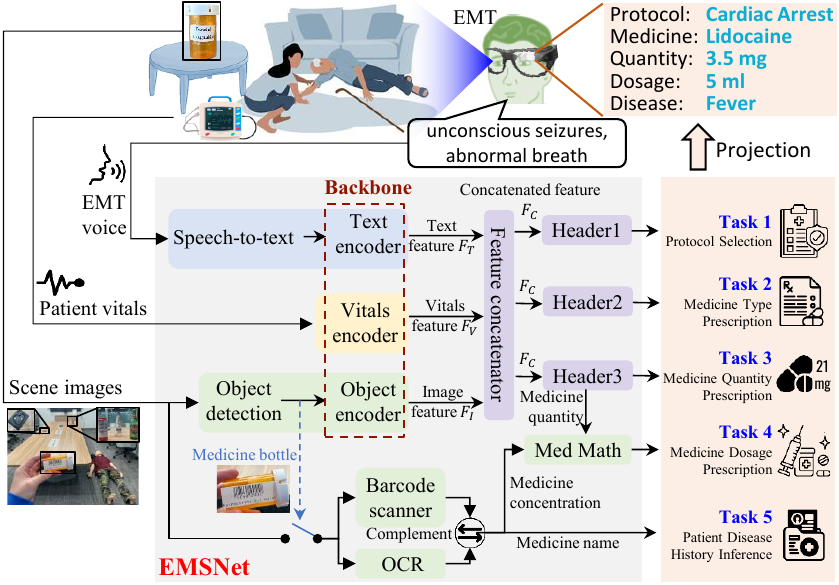}
    \caption{Overview of EMSNet for EMS}
    \vspace{-3.5mm}
    \label{fig:model_overview}
\end{figure}

Figure~\ref{fig:model_overview} illustrates the modular design of EMSNet, the first on-glass multimodal multitask model for EMS. When an EMT wearing smart glasses encounters a collapsed patient at the EMS scene, they verbally report the patient's symptoms ``unconscious seizures, abnormal breath''. The microphone on the smart glasses captures voice, processed by the~\ul{text module}, i.e., speech-to-text module and text encoder. The patient's numeric time-series vitals, obtained from medical equipment, are processed by the~\ul{vitals module}, i.e., vitals encoder. The smart glasses camera captures scene images, analyzed by the~\ul{image module}, i.e., object detection and object encoder. We call the three encoder modules~\ul{backbone} as their output features ($F_T$, $F_V$, and $F_I$) are concatenated, and then form a unified feature representation $F_C$. $F_C\in\mathcal{R}^{|F_T|+|F_V|+|F_I|}$ serves as input to three distinct header modules producing recommendations for three tasks: Task 1) protocol selection recommendation, Task 2) medicine type prescription for recommending which type of medicine to administer, and Task 3) medicine quantity prescription for recommending specific medication quantities. 
If the object detection module locates a medicine bottle from the scene, the original scene image is processed by the OCR and barcode scanner for medicine name and concentration from the label. The Med-Math module then takes the quantity from header 3, using concentration to determine Task 4) dosage prescription. Medicine names undergo a dictionary check to infer Task 5) patient's disease history. Both OCR and barcode scanner are needed because we find medicine labels may not always contain both names and barcodes. Complementing the outputs from both pipelines ensures EMSGlass can always extract the desired information. Finally, recommendations of all five tasks are continuously updated on the display of the glasses. Table~\ref{tab:emsfoundation_components} lists model candidates for all component modules of EMSNet. Before we dive into the design details of these module candidates, we first describe how we prepare the multimodal EMS dataset to train and test EMSNet.

\begin{table}[h]
\centering
\caption{Model candidates of each component in EMSFoundation.}
\vspace{-8pt}
\label{tab:emsfoundation_components}

\scalebox{0.69}{

\begin{tabular}{l|l}
\hline
Backbone Text encoder(3)   & TinyBERT, MobileBERT, BERTBase                         \\
Backbone Vitals encoder(3) & RNN, LSTM, GRU                                         \\
Backbone Image encoder(1)  & fully connect (FC)                                     \\
Headers(1)                 & Feature concatenator + header1 + header2 + header3     \\
\hline
Speech-to-text(4)          & Whisper-tiny, -base, -small, -medium \\
Object detection(2)        & YOLO11n, YOLO11x                                            \\
\hline
OCR(4)                     & EasyOCR, TesseractOCR, PaddleOCR, CRNN                 \\
Barcode scanner(1)            & ML Kit                                                        \\
Med Math(1)                & A division operator                                    \\ \hline
\end{tabular}

}
\vspace{-5mm}
\end{table}

\subsection{Multimodal multitask data processor}
\label{sec:multimodal_processor_design}

\begin{figure}
    \centering
    \includegraphics[width=\linewidth]{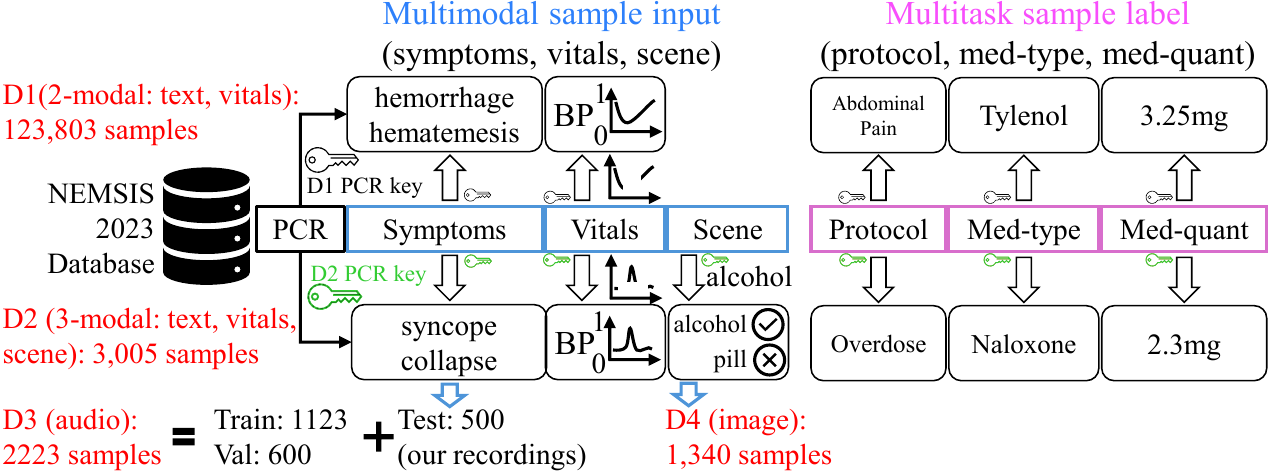}
    \vspace{-6mm}
    \caption{Pipeline of the multimodal data processor to prepare 4 datasets for training our EMSFoundation model: D1(2-modal: text, vitals), D2 (3-modal: text, vitals, scene), D3(audio), and D4(image).}
    \vspace{-8pt}
    \label{fig:data_processor}

\end{figure}

Figure~\ref{fig:data_processor} illustrates the data processing pipeline used to prepare four multimodal datasets (D1-D4) to train and test EMSNet. This pipeline is designed as a general plug-n-play tool for preparing multimodal datasets from NEMSIS~\cite{nemsis2015}, a publicly available (upon request) key-value tabular database that includes all real-world national emergency event reports in the US. Notably, \textit{our data processor is the first tool of its kind to prepare multimodal EMS datasets}. The NEMSIS database uses a unique 9-digit ``PCR key'' as each emergency event ID. In this paper, we utilize the NEMSIS 2023 database, which includes over 54 million EMS events recorded by 14,369 EMS agencies across 54 states and territories in 2023.

\textbf{D1(2-modal: text,vitals)}: We get D1's PCR keys from NEMSIS, extracting events that contain symptom data (primary symptom, primary impression, associate symptom, and secondary impression). Since EMTs typically use symptoms to describe patients' medical status~\cite{EMSAssist2023MobiSys}, we concatenate the four symptom texts into a single sentence as the text-mode input to EMSNet. Similarly, we use the same PCR keys for time-series vitals (BP-blood pressure, HR-heart rate, PO-pulse oximetry, RR-respiratory rate, CO$_2$-end-tidal carbon dioxide, BG-blood glucose) inputs. The extracted vitals are preprocessed with a few steps detailed in Appendix~\ref{sec:multimodal_processor_implement}, e.g., outlier removal, padding missing values, and cross-sample normalization.
On the right, we use the same keys to extract labels for each sample: protocol (abdominal pain) for Task 1 (protocol selection), medicine type (naloxone) for Task 2 (medicine prescription), and quantity (3.25 mg) for Task 3 (medicine quantity). \textit{This data processor is the first to prepare and enable the Task 2 and 3 for EMS}. The final dataset is a 2-modal (text,vitals) set with 123,803 samples.

\textbf{D2(3-modal: text, vitals,scene)}: Similarly, we extract D2's text and vitals with D2's PCR Keys, a disjoint set of D1 corresponding to events with 3-modal input: text, vitals, and scene data. We also extract the EMS scene information indicating the existence of alcohol and pills. The scene information is one-hot encoded, giving a 3-modal dataset D2(text, vitals, scene) with 3,005 samples.

\textbf{D3(audio)}: To train an accurate speech-to-text model, we use the open-sourced audio dataset (1123 for training and 600 for validation) from EMSAssist~\cite{EMSAssist2023MobiSys}. For testing, with 50 randomly sampled symptom texts from D2, five users recorded 50 audio samples using HyperX Solocast microphone~\cite{HyperXSolocast2024} and Google Glass~\cite{GoogleEEII2024}, giving a set of 500 audio samples. 

\textbf{D4(image)}: We place 3 types of EMS objects (alcohol bottles, pills, and medicine bottles) on a table in 5 different rooms shown in Figure~\ref{fig:user_study_room} in Appendix~\ref{sec:user_study_room}: (a) bedroom, (b) living room, (c) student office, (d) conference room, and (e) user study room. We recorded videos of objects using Google Glass~\cite{GoogleEEII2024} in these 5 rooms. From these video recordings, we sample 1,240 images from rooms (a)-(d) for training (908) and validation (232) sets, and 200 images from room (e) for testing. The final image dataset contains 1,340 samples for training and testing EMS-specific object detection models. 

\vspace{-5pt}
\subsection{Backbone and headers}
\label{sec:backbone_headers}
\vspace{-3pt}

As shown in Figure~\ref{fig:model_overview} and Table~\ref{tab:emsfoundation_components}, the backbone includes three encoder models: 1) a symptom text encoder, 2) a vitals encoder, and 3) an image encoder. We use TinyBERT, MobileBERT, and BERTBase for the symptom text encoder as they have demonstrated success in encoding text data, including EMS texts~\cite{CognitiveEMS2024IoTDI,EMSAssist2023MobiSys,devlin2019bert}. RNN, LSTM, and GRU are employed for time-series vitals data. A fully-connect (FC) operator is applied to encode the one-hot vector from the object detection model. To fuse multimodal features, we use a feature concatenator to concatenate three encoders' output vectors into one longer vector. We also tried other feature fusion methods, including dot product, weighted sum, weighted concatenation, and attention. We ultimately choose concatenation due to its higher accuracy on the multimodal EMS dataset. Headers 1 and 2 are standard one-layer classification headers while header 3 is a standard regression header. For simplicity, in this paper, we call the collection of the feature concatenator and three headers as headers. 

We use D1 (2-modal: text, vitals) and D2 (3-modal: text, vitals, scene) to train the multimodal backbone with PyTorch~\cite{pytorch2019}. During training, we employ top 1/3/5 categorical loss for headers 1 and 2, while mean square error (mse), pearson coefficient (pearsonr$\in [0,1]$)~\cite{Pearsonr}, and spearmanr coefficient (spearmanr$\in [0,1]$)~\cite{Pearsonr} for the header 3.

\textbf{Multimodal training with progressive modality integration (PMI)}: Multimodal training generally suffers from the imbalanced dataset size among different modalities. For example, the 2-modal D1 has a much larger dataset size (123,803) than 3-modal D2 (3005). Although using D1 alone may get a performant 2-modal model due to sufficient amount of 2-modal data samples, 
directly training the 3-modal model on D2 from scratch will significantly degrade the 3-modal accuracy. To address this, we propose progressive modality integration (PMI)~\cite{shankar2022progressivefusionmultimodalintegration}. Instead of training the 3-modal model from scratch, we leverage the pre-trained 2-modal model. Specifically, in each epoch of 3-modal training, instead of passing the batched samples to a newly initialized 3-modal model, we feed the 2-modal parts (text, vitals) of the 3-modal sample into the trained 2-modal model to get feature outputs ($F_C \in \mathcal{R}^{|F_T|+|F_V|}$), while passing the scene part of the batched samples to the newly initialized scene encoder to get $F_I$. We combine it with the 2-modal's $F_C$ for a new $F_C$. Since the 2-modal $F_C$ is an order of magnitude larger than $F_I$, it retains most of the knowledge learned from D1 while integrating additional scene information from D2. Our evaluation confirms that PMI effectively enables competent 3-modal backbone models.

\subsection{Speech-to-text and object detection}
\subsubsection{Speech-to-text}

\begin{figure}[t]
    \centering
    \begin{subfigure}{0.44\linewidth} 
        \centering
        \includegraphics[width=\linewidth]{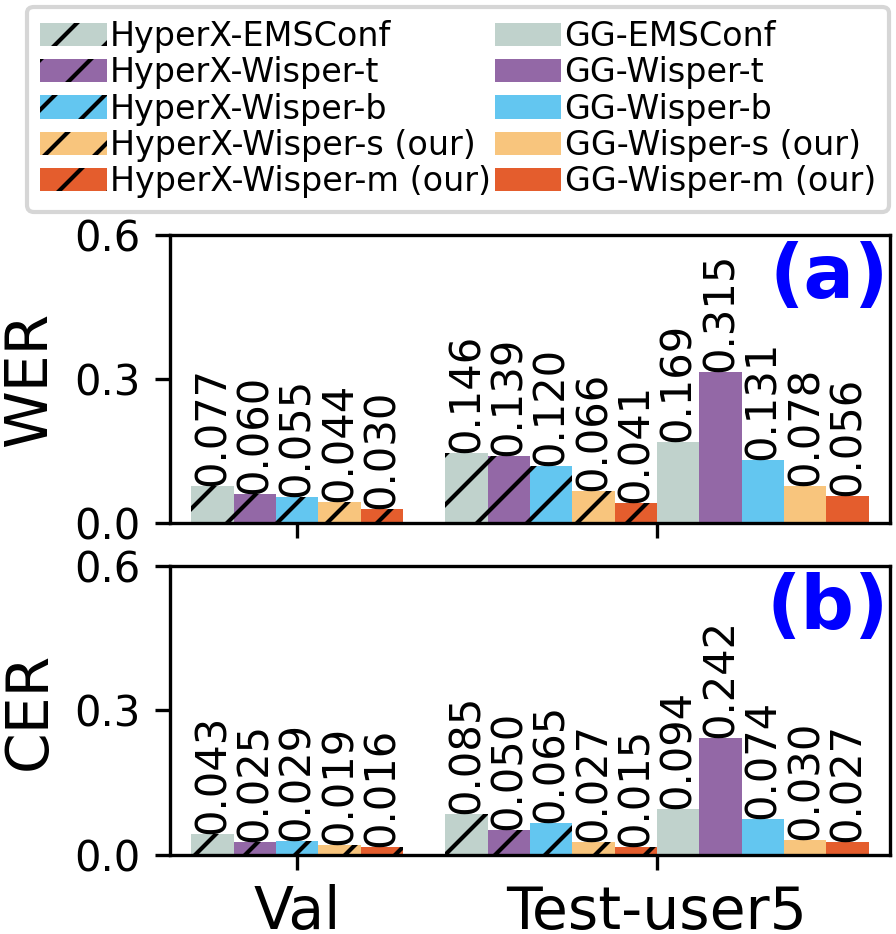}
    \end{subfigure}%
    \hfill
    \begin{subfigure}{0.47\linewidth} 
        \centering
        \includegraphics[width=\linewidth]{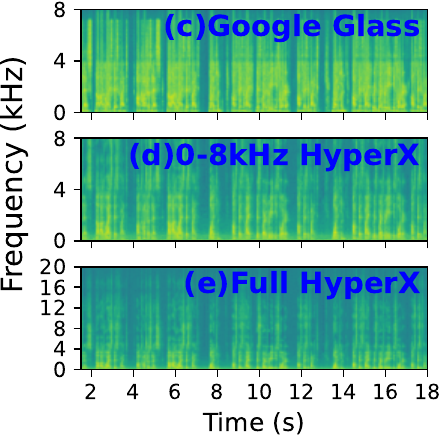}
    \end{subfigure}
    \caption{(a)-(b): Whisper-s and Whisper-m achieve lower error rates and improved generalization across user accents and microphones. (c)-(e): Spectrograms show high-frequency signal loss in Google Glass due to an 8kHz cutoff.}
    \label{fig:sr_wer_cer_motivate}
\end{figure}

State-of-the-art EMS assistants~\cite{PreumTowardsICCPS2018,PreumCognitiveEMSSigbed2019,ShuBehaviorIROS2019} rely on cloud-based speech-to-text services like Google Speech-to-Text, which require stable internet access—often unavailable in disaster-stricken areas. To address this, EMSAssist~\cite{EMSAssist2023MobiSys} and CognitiveEMS~\cite{CognitiveEMS2024IoTDI} fine-tuned small models (e.g., EMSConformer, Whisper-tiny, and Whisper-base) with custom audio datasets for reliable on-device transcription. However, these models exhibit poor generalization across user accents and microphone hardware. In Figure~\ref{fig:sr_wer_cer_motivate}(a-b), accuracy degradation is evident in dataset D3(audio), where HyperX-recorded validation data achieves lower word error rates (WER) and character error rates (CER) than user5's test set. Additionally, pervasive microphone frequency cutoffs on mobile devices~\cite{NFON2024, FlexRadio2024, Audio-Technica2024, asmar2019phone4khz}, such as Google Glass’s 8kHz limitation, cause substantial speech information loss. Spectrograms in Figure~\ref{fig:sr_wer_cer_motivate}(c-e) confirms this, showing lost high-frequency signals in Google Glass recordings. Simultaneously recorded audio tests reveal Whisper-t’s WER rises from 0.139 (HyperX) to 0.315 (Google Glass).

These challenges stem from data distribution shift~\cite{HwangDistributionECCV2022,ChipDataDistributionOnline2024}, where real-world inputs—affected by diverse accents and microphones—differ from training data, yet the labels remain unchanged. The root cause is the small size of existing models, all under 100 million parameters (EMSConformer: 10m, Whisper-tiny: 74m, Whisper-base: 74m), limiting both learning and generalization. Empirical evidence~\cite{kaplan2020scalinglawsneurallanguage,alabdulmohsin2022revisitingscalinglaws} supports this approach, showing that larger models achieve lower training loss. Specifically, we propose training Whisper-small (242m) and Whisper-medium (764m) on dataset D3(audio) to enhance generalization (as shown in Figure~\ref{fig:sr_wer_cer_motivate}). Here we select one order of magnitude higher than the SOTA to demonstrate the effectiveness of EMSWhisper's idea in enhancing generalization capability. Further explorations of alternative scaling options could be considered future work.

\subsubsection{Object detection}
\label{sec:od_model_design}

\begin{figure}[h]
    \includegraphics[width=1\linewidth]{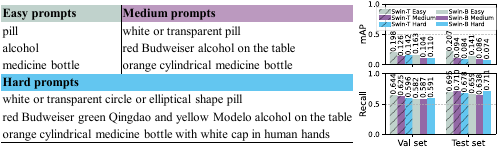}
    \vspace{-15pt}
\caption{With different text prompts(left), we evaluate (right) Grounding Dino on the collected EMS scene image dataset containing pills, alcohol, and medicine bottles.}
    \label{fig:gd_prompts_map_recall_vertical}
\end{figure}

To reduce the manual effort required for data annotation and model training in developing an EMS-specific object detector, we assessed the SOTA open-set detection model Grounding DINO\cite{liu2024groundingdinomarryingdino} (GD) using prompts of varying difficulty for humans to come up with (See Figure~\ref{fig:gd_prompts_map_recall_vertical} (left): easy, medium, and hard).

Figure~\ref{fig:gd_prompts_map_recall_vertical} (right) shows the mean average precision (mAP) and recall performance for two GD models (Swin-T: 172m, Swin-B: 341m), where higher mAP and recall indicate better performance. The observed disparity between low mAP (<0.2) and high recall (>0.55) suggests that open-set object detection models like GD generate excessive false positives, limiting its suitability in generalizing EMS objects detection, where precision is crucial. Therefore, a dedicated EMS-specific detector remains necessary.

Close-set detection models (e.g., YOLO11~\cite{YOLO11Ultralytics2024}), in contrast to the aforementioned open-set Grounding Dino (GD) model, require manual dataset annotation and model training when generalized onto a new domain, e.g., EMS. To relieve the demanding annotation workloads, we propose \textit{human-in-the-loop (HITL) annotation adjustment}, leveraging the open-set strength of the GD into the human annotation process. There are two rounds of annotations: in the first round, we feed unlabeled EMS scene images D4(image) to GD to get bounding box labels. In the second round, instead of annotating from scratch, we adjust GD's annotations from the first round (e.g., relabel GD's incorrect annotations). After two rounds, we fine-tune the close-set YOLO11. When compared to conventional manual annotation without GD's auto-labeling process, our HITL annotation adjustment saves the annotation time by exploiting the correct annotations from GD and leaves the main manual efforts in GD's incorrect or missed annotations. From our experiences, this HITL annotation adjustment method decreases the annotation time by half, e.g., on average, from 10 to 5 minutes for every 100 images.  

\subsection{OCR, barcode scanner, and med math}
\label{sec:ocr_barcode_medmath}

\begin{figure}
    \centering
    \includegraphics[width=\linewidth]{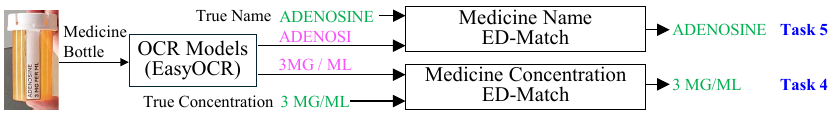}
    \caption{Pipeline of the OCR workflow using edit distance matching (ED-Match) with ground truth medicine name and concentration}
    \label{fig:ocr_design}
\end{figure}

Once a medicine bottle object is detected in the scene image, EMSNet in Figure~\ref{fig:model_overview} uses optical character recognition (OCR) and barcode scanner to extract medicine details from the original image. As shown in Figure \ref{fig:ocr_design}, we use Easy-OCR~\cite{EasyOCR2023} to extract medicine names and concentrations on the labels of the medicine bottle. To filter out wrong outputs, we employ an edit distance (ED)-based post-processing module for matching all outputs with a predefined true list of EMS medicines and corresponding concentration. In cases where the text information on the medicine bottle label is missing but the barcode is available, we use ML Kit~\cite{MLKit} to build a barcode scanner to complement the OCR by extracting the medicine names and concentration from the barcode. As exemplified in Section~\ref{sec:object_detection_ocr_background}, OCR/scanner's medicine concentration output, combined with the medicine quantity prescription from header 3, is input to med math--a division operator--to compute the dosage of the medicine solution for task 4. The medicine name output is mapped to a list of 82 common EMS diseases for accomplishing task 5.


\section{EMSServe design}
\label{sec:emsserve_design}

\textit{EMSServe is the first multimodal model serving framework for EMS scenarios with multimodal data arriving at different times}. Before we dive into the design details, we would like to clarify: as in Section~\ref{sec:system_design}, a ``module'' (e.g., text module, vitals module) refers to EMSNet components; in contrast, a ``text-vital'' model includes both text and vitals modules with additional headers for accomplishing EMS tasks.

\begin{figure}[h]
    \includegraphics[width=\linewidth]{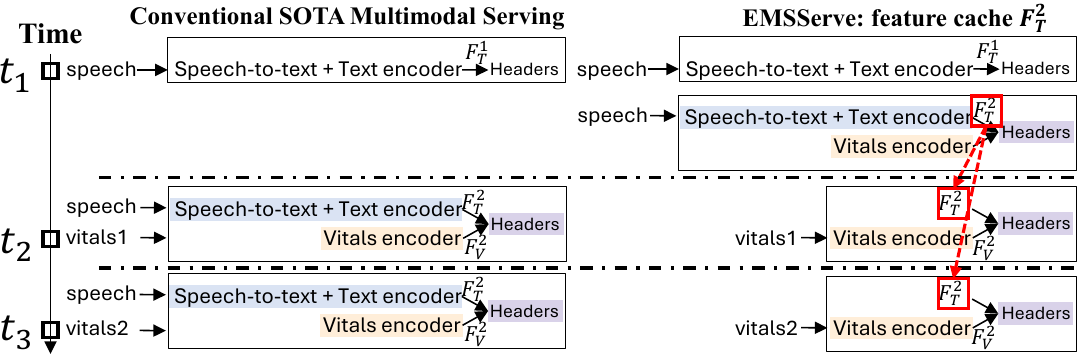}
\caption{When compared to state-of-the-art multimodal serving systems, EMSServe's modality-aware cache minimizes redundant computations in the text submodule by accounting for varying data arrival times across modalities.}
    \label{fig:emsserve_idea}
\end{figure}

\subsection{Motivations and insights}

\begin{figure*}[t] 
    \centering
    \begin{subfigure}{0.807\linewidth} 
        \centering
        \includegraphics[width=\linewidth]{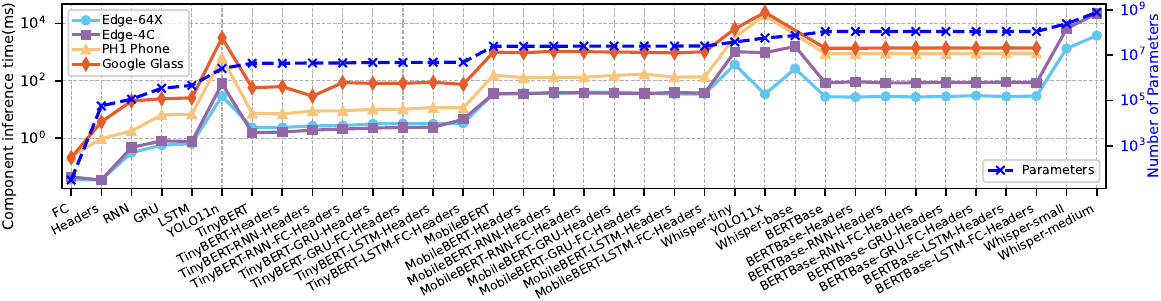}
    \end{subfigure}%
    \hspace{0.002\linewidth}
    \begin{subfigure}{0.186\linewidth} 
        \centering
        \includegraphics[width=\linewidth]{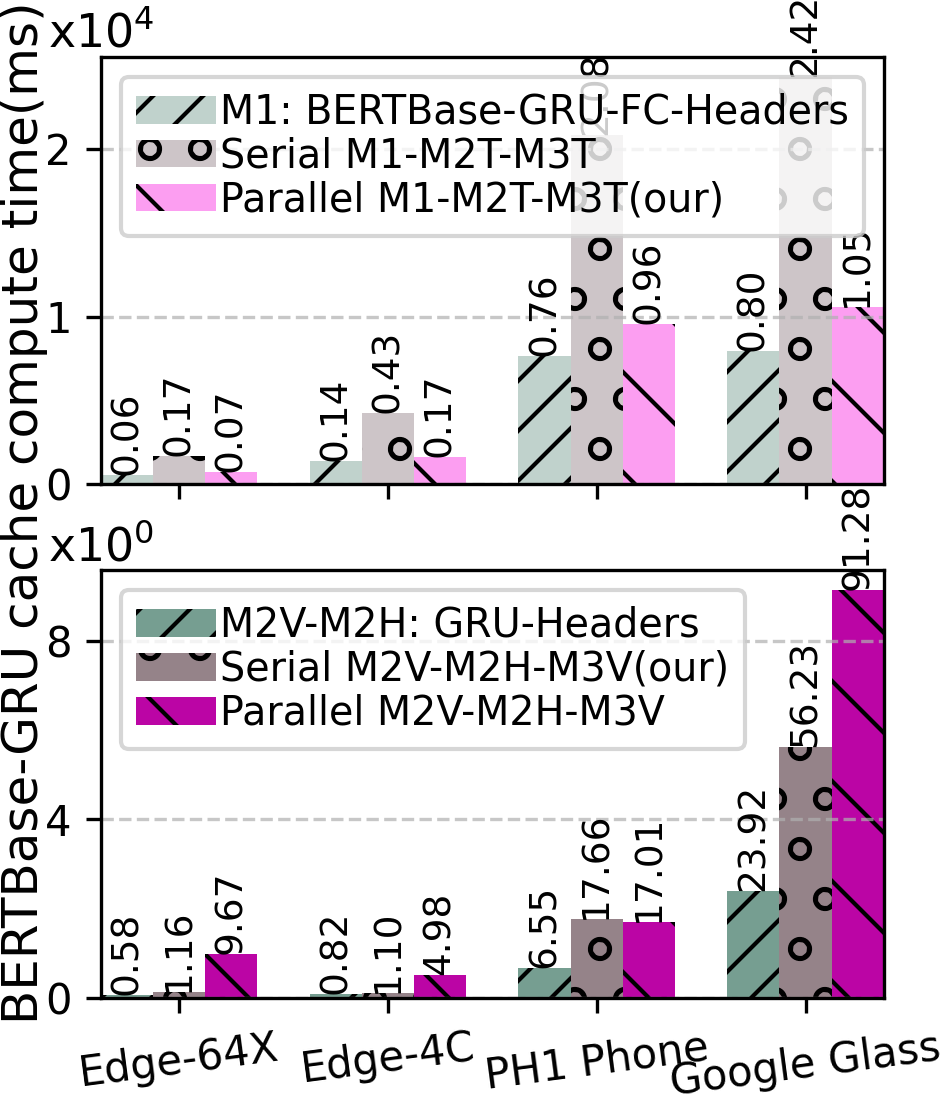}
    \end{subfigure}
    \caption{(left)--Inference time of different multimodal foundation model components on 4 hardware platforms: edge server with 64 Intel Xeon CPU (Edge-64X), edge server with 4 Intel Core CPU (Edge-4C), PH1 phone, and Google Glass; (right)--On both mobile devices and edge servers, the parallelism is advantageous to compute feature cache for compute-expensive text submodules(right top) of multimodal EMSFoundation model whereas serial execution is advantageous to compute feature cache for inexpensive vitals submodules(right bottom).}
    \vspace{-10pt}
    \label{fig:backbone_components_inference_time}
\end{figure*}


\textbf{Insight 1 -- Avoid redundant text module inference}: Figure~\ref{fig:different_time_arrival} illustrates an EMS scenario where an EMT's speech describing symptoms and a patient’s vitals arrive at the smart glasses at different times. In conventional multimodal frameworks like PyTorch (Figure~\ref{fig:emsserve_idea}, left), the speech is first processed by a text model (i.e., speech-to-text + text encoder + headers) at \(t_1\). When the first vitals data (vitals1) arrives at \(t_2\), the system recomputes the text submodule output by feeding both speech and vitals1 into a text-vital model. This process repeats when additional vitals (vitals2) arrive at \(t_3\), leading to the redundant inference of the text module (speech-to-text + text encoder). Real-world EMS data (NEMSIS 2023~\cite{nemsis2015}) records up to 30 vitals per event, potentially causing the text submodule to run up to 30 times.


\begin{table}[t!]
\centering
\caption{Hardware and software platform specifications}
\label{tab:hardware}
\scalebox{0.7}{
\begin{tabular}{ll}
\hline
\multicolumn{2}{l}{\textbf{Edge-64X}}                          \\
CPU                & 64 x Intel Xeon Silver 4314      \\
RAM                & 64 GB                            \\
OS                 & Ubuntu 22.04                     \\
Torch version      & PyTorch 2.3                      \\ \hline
\multicolumn{2}{l}{\textbf{Edge-4C}}                           \\
CPU                & 4 x Intel Core i7 7567U          \\
RAM                & 16 GB                            \\
OS                 & Ubuntu 22.04                     \\
Torch version      & PyTorch 2.3                      \\ \hline
\multicolumn{2}{l}{\textbf{PH1 Phone}}                         \\
SoC                & Qualcomm Snapdragon 835          \\
OS                 & Android 9                        \\
Torch version      & PyTorch Android Lite 2.1         \\ \hline
\multicolumn{2}{l}{\textbf{Google Glass Enterprise Edition 2}} \\
SoC                & Qualcomm XR1                     \\
OS                 & Android 8                        \\
Torch version      & PyTorch Android Lite 2.1         \\ \hline
\end{tabular}
}
\end{table}

\textbf{Insight 2 -- Text modules dominate EMSNet inference time}: Figure~\ref{fig:backbone_components_inference_time} shows that text modules, including the speech-to-text and text encoder, are the primary inference bottleneck across four hardware platforms: Edge-64X, Edge-4C, PH1 Phone, and Google Glass Enterprise Edition 2 (hardware details in Table~\ref{tab:hardware}). Whisper variants, the speech-to-text models, require significantly more inference time than other components. Whisper-small and Whisper-medium, the largest models, are the slowest on edge servers and are entirely unusable on PH1 and Google Glass due to memory limitations. Additionally, text encoders further exacerbate latency. For instance, components incorporating BERTBase exhibit inference times nearly identical to BERTBase itself, with similar trends for TinyBERT and MobileBERT. In contrast, vitals submodules (RNN, GRU, LSTM) and headers require substantially less processing time, underscoring the inefficiency challenge of repeated text submodule computations highlighted in Insight 1.

\textbf{Key idea of EMSServe--feature cache}: To eliminate redundant text submodule inference, EMSServe employs feature cache (Figure~\ref{fig:emsserve_idea}, right). When voice data arrives, instead of only running the text model, we simultaneously compute and cache the text module output $F_T^2$ in the text-vital model. When vitals1 arrives at \( t_2 \), unlike conventional frameworks that rerun the costly text submodule, EMSServe reuses the cached \( F_T^2 \) and processes only the vitals module (vitals encoder) to obtain \( F_V^2 \), which significantly reduces inference costs. The cached \( F_T^2 \) and newly computed \( F_V^2 \) are then concatenated as inputs to model headers. By addressing asynchronous EMS data arrival times, EMSServe mitigates redundant and costly submodule computation to improve inference efficiency.

\begin{figure*}[h]
    \centering
    \includegraphics[width=\textwidth]{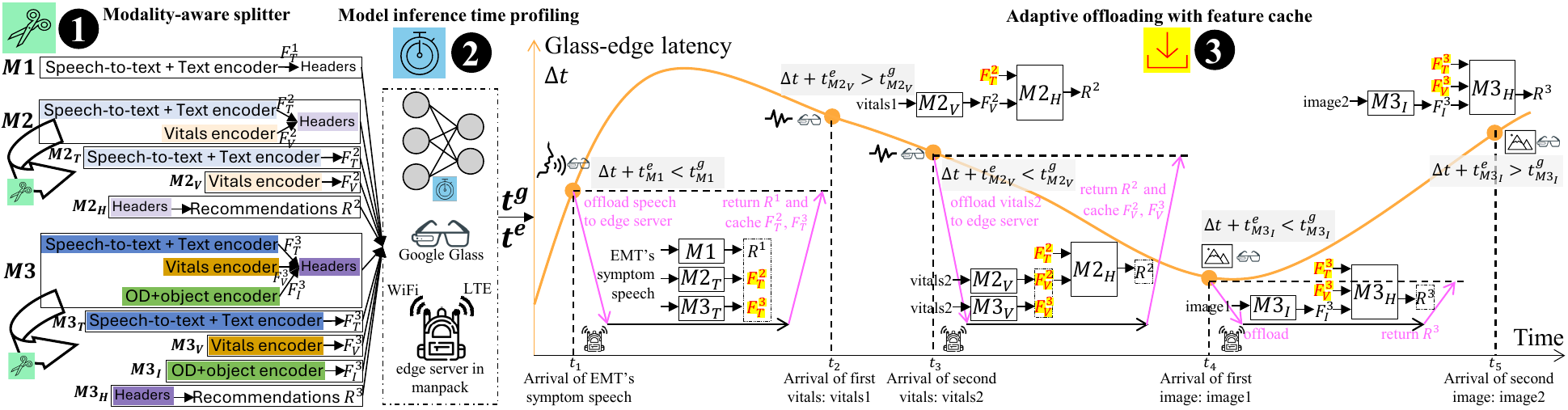}
    \caption{Overview of EMSServe, the first serving framework for multimodal model inference in EMS scenarios.}
    \vspace{-12pt}
    \label{fig:emsserve_oview}
\end{figure*}

 \textbf{Insight 3 --- inference latency varies rapidly across hardware}: Figure~\ref{fig:backbone_components_inference_time} shows that inference latency differs significantly across hardware. Google Glass, with limited computing power, presents the highest latency, while Edge-64X, with ample resources, is the fastest. For instance, YOLO11n inference takes 3.2s on Google Glass, 0.7s on PH1, 0.08s on Edge-4C, and just 0.03s on Edge-64X. This disparity implies an opportunity for latency reduction through edge-assisted workload offloading. Offloading YOLO11n from Google Glass to Edge-64X achieves over 100× speedup. EMSServe exploits this by offloading high-latency tasks, such as text module inference and object detection, to edge servers.

\subsection{EMSServe design}

Building on our insights, Figure~\ref{fig:emsserve_oview} abstracts EMSServe, the first serving framework for multimodal model inference in EMS. EMSServe comprises three key components: \darkcircled{1} a modality-aware splitter that decomposes multimodal models into single-mode modules, \darkcircled{2} model inference time profiling to measure inference latency on the smart glasses and the edge server, and \darkcircled{3} adaptive offloading with feature caching to minimize latency in dynamic EMS scenarios. While we evaluate EMSGlass on PH1 and Edge-64X in Figure~\ref{fig:backbone_components_inference_time} for generality, our primary setup uses Google Glass as the mobile device and a manpack-mounted Edge-4C as the edge server.

\subsubsection{Modality-aware splitter} As shown in Figure~\ref{fig:emsserve_oview}, the modality-aware splitter decomposes the multimodal model $M2$ ($M3$) into single-mode modules $M2_T$ and $M2_V$ ($M3_T$, $M3_V$, and $M3_I$), enabling precomputation and cache of the text module output features $F_T^2$ ($F_T^3$) before vitals data arrive.

\subsubsection{Model inference time profiling} After splitting, we profile all models and modules to obtain their inference times $t^g$ and $t^e$ on Google Glass and Edge-4C. Figure~\ref{fig:backbone_components_inference_time} details the profiling results. Both modality-aware splitting and profiling are one-time offline efforts, providing $t^g$ and $t^e$ as inputs for EMSServe’s real-time multimodal request serving.

\subsubsection{Adaptive offloading with feature cache} \label{sec:adaptive_offloading}
Figure~\ref{fig:emsserve_oview} illustrates how EMSServe dynamically serves multimodal inference in an EMS scenario where symptom speech, vitals, and image data arrive asynchronously. The EMT wearing Google Glass and the EMT carrying the manpack-mounted Edge-4C move independently, causing bandwidth (BW) fluctuations in Glass-edge communication. To monitor this variation in real-time, we implement a lightweight heartbeat monitor on the smart glasses, periodically (e.g., every second) measuring the file transmission bandwidth $\Delta t = BW * file size$. Unlike RTT, $\Delta t$ represents the actual file transfer time. Based on this measurement, EMSServe optimizes offloading decisions:  

\begin{itemize}
    \item Low Latency ($\Delta t$ is low): offload compute-heavy tasks (e.g., text submodule inference, object detection) to the edge server for its higher processing power.
    \item High Latency ($\Delta t$ is high): avoid offloading, as communication delay outweighs computation benefits, making on-device inference preferable.  
\end{itemize}

To make these decisions efficiently in real-world EMS scenarios with high mobility, EMSServe employs adaptive offloading with a feature cache, executed in three key steps:

1) When the EMT's symptom speech arrives at time \( t_1 \), EMSServe evaluates the real-time Glass-edge latency (\( \Delta t \)) and the profiled inference time of model \( M1 \) on the edge server (\( t_{M1}^e \)). Since the total offloading time (\( \Delta t + t_{M1}^e \)) is lower than the inference time on Google Glass (\( t_{M1}^g \)), the system offloads the speech processing to the edge server. The edge server returns recommendations \( R^1 \) and precomputed text module feature cache \( F_T^2 \) and \( F_T^3 \) from models \( M2_T \) and \( M3_T \), respectively, reducing redundant inference for future multimodal tasks.  

2) Upon the arrival of the first vitals data at time \( t_2 \), on-glass inference is chosen because the estimated offloading delay (\( \Delta t + t^e_{M2_V} \)) exceeds the inference time on Google Glass (\( t^g_{M2_V} \)). The vitals data is processed by the pre-split vitals module \( M2_V \), generating \( F^2_V \). Using the cached text module output \( F_T^2 \), EMSServe efficiently computes the recommendation \( R^2 \) on Google Glass without running the full multimodal model. When a second vitals data point arrives at \( t_3 \), the system instead offloads processing to the edge server (\( \Delta t + t^e_{M2_V} < t^g_{M2_V} \)). In parallel, the edge server also processes the vitals module \( M3_V \) to generate feature cache \( F^3_V \), which is returned alongside \( R^2 \) and \( F^2_V \), ensuring cache consistency.  

3) Similarly, for EMS scene images, EMSServe offloads processing when the offloading delay is lower than the on-glass inference time. When the first image arrives at \( t_4 \), it is offloaded to the edge server, where the image submodule \( M3_I \) extracts feature \( F^3_I \). This is combined with cached features \( F^3_T \) and \( F^3_V \) precomputed in previous steps to generate recommendation \( R^3 \), which is sent back to Google Glass. When the second image arrives at \( t_5 \), the on-glass inference is chosen due to high offloading latency. With previously cached text and vitals features, the system processes only the image submodule \( M3_I \), avoiding redundant inference of other modalities.

\textbf{Parallel cache computation}: We utilize multiple parallel threads to reduce cache computation overheads. Figure~\ref{fig:backbone_components_inference_time} (right) shows the compute time for the text feature cache (step 1) and vitals cache (step 2) across all hardware platforms, including Google Glass. For the text module cache (top), parallel computation takes nearly the same time as running the text model \( M1 \) alone, effectively hiding the text feature cache computation from users since \( M1 \) must be computed regardless. However, for the vitals module (bottom), which inherently takes much less time to compute by up to four orders of magnitude, launching multiple parallel threads often incurs higher costs than serial execution. Hence, we adopt serial feature cache computations for vitals.

\textbf{Fault tolerance for edge server crashes}: To address potential failures of manpack edge server, such as battery depletion, we implement a dual-perspective fault tolerance mechanism. First, at each step, when the edge server transmits both inference results and recommendation results to the smart glasses, it also returns the computed feature cache. This ensures that the cache on the smart glasses is never outdated by more than one step. Second, if the edge server crashes during feature cache computation, the smart glasses seamlessly switch to on-device inference using the current step's data. This design maintains real-time recommendation updates with minimal disruption.

\begin{figure}
    \includegraphics[width=0.99\linewidth]{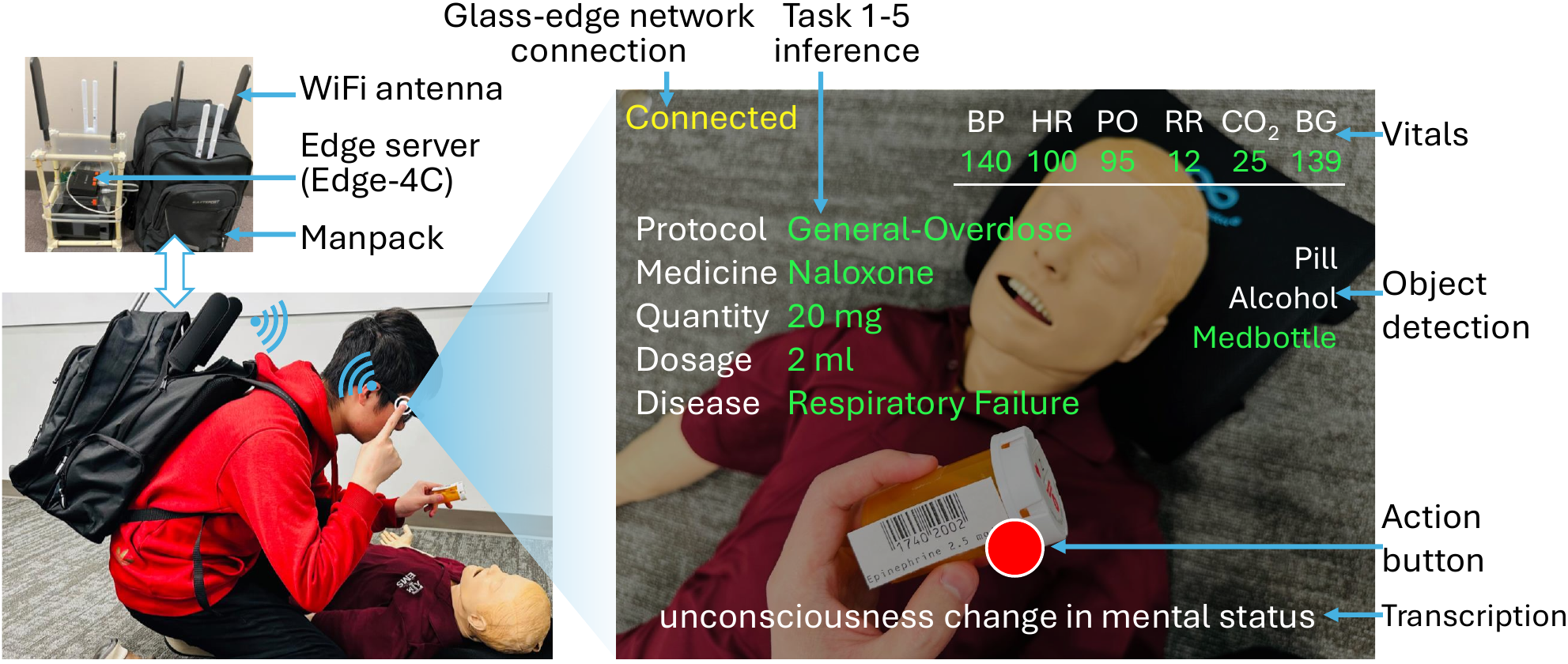}
    \vspace{-10pt}
\caption{User interaction for networked Google Glass and Edge-4C (intel NUC edge server with WiFi) packed into a manpack.}
    \vspace{-10pt}
    \label{fig:ui_design}
\end{figure}

\subsubsection{User interaction design and implementation}  
Seamless UI design is crucial for EMTs to adopt networked smart glasses, given their mobility and resource constraints (Section~\ref{sec:intro}). Figure~\ref{fig:ui_design} illustrates the on-display UI: the left screen continuously updates recommendations for EMS tasks while tapping the frame (right bottom) triggers EMSGlass to capture EMTs' symptoms speech, patients' vitals, and scene images. Synchronously arrived data is offloaded adaptively (Section~\ref{sec:adaptive_offloading}) via WiFi to the Edge-4C server carried in a manpack, ensuring low-latency inference and real-time recommendation on-display updates in high mobility EMS scenarios. 
EMSGlass is implemented as an Android app on Google Glass Enterprise Edition 2~\cite{glassSpecs2023}, comprising over 3,000 lines of Java. EMSServe, built on Torch Serve~\cite{torchserve}, runs on Edge-4C with ~2,000 lines of Python. HTTPS communication enables multimodal inference offloading, ensuring efficient processing despite smart glasses' resource limitations.

\section{EMSGlass Evaluation}
\label{sec:evaluation}

\begin{table*}[]
\centering
\caption{Compare the 2-modal EMSNet with state-of-the-art (SOTA) unimodal models in tasks 1-3 on the 2-modal dataset D1 (text, vitals).}
\label{tab:multimodal_unimodal_d1}
\scalebox{0.56}{
\begin{tabular}{lllcccccccccccc}
\hline
 &
  \multicolumn{2}{c}{Backbone: T/V} &
  \multicolumn{4}{c}{Task1-Protocol Selection (top-1/3/5)} &
  \multicolumn{4}{c}{Task2-Medicine Type Prescription (top-1/3/5)} &
  \multicolumn{4}{c}{Task3-Medicine Quant. Prescription (mse/pearsonr/spearmanr)} \\
\multirow{-2}{*}{} &
  T &
  V &
  P &
  P-M &
  P-Q &
  P-M-Q &
  M &
  P-M &
  M-Q &
  P-M-Q &
  Q &
  P-Q &
  M-Q &
  P-M-Q \\ \hline
 &
   &
  LSTM &
  {\color[HTML]{9B9B9B} 0.43/0.66/0.76} &
  {\color[HTML]{9B9B9B} 0.38/0.60/0.71} &
  {\color[HTML]{9B9B9B} 0.38/0.61/0.71} &
  {\color[HTML]{9B9B9B} 0.38/0.61/0.71} &
  {\color[HTML]{9B9B9B} 0.50/0.77/0.87} &
  {\color[HTML]{9B9B9B} 0.49/0.77/0.87} &
  {\color[HTML]{9B9B9B} 0.50/0.77/0.87} &
  {\color[HTML]{9B9B9B} 0.49/0.76/0.87} &
  {\color[HTML]{9B9B9B} 2.63/0.21/0.22} &
  {\color[HTML]{9B9B9B} 2.73/0.20/0.20} &
  {\color[HTML]{9B9B9B} 2.63/0.21/0.22} &
  {\color[HTML]{9B9B9B} 2.72/0.20/0.20} \\
 &
   &
  RNN &
  {\color[HTML]{9B9B9B} 0.44/0.68/0.77} &
  {\color[HTML]{9B9B9B} 0.40/0.63/0.74} &
  {\color[HTML]{9B9B9B} 0.39/0.63/0.74} &
  {\color[HTML]{9B9B9B} 0.39/0.63/0.74} &
  {\color[HTML]{9B9B9B} 0.51/0.80/0.90} &
  {\color[HTML]{9B9B9B} 0.50/0.80/0.89} &
  {\color[HTML]{9B9B9B} 0.50/0.79/0.89} &
  {\color[HTML]{9B9B9B} 0.50/0.79/0.89} &
  {\color[HTML]{9B9B9B} 2.64/0.20/0.20} &
  {\color[HTML]{9B9B9B} 2.71/0.21/0.23} &
  {\color[HTML]{9B9B9B} 2.62/0.22/0.23} &
  {\color[HTML]{9B9B9B} 2.71/0.21/0.23} \\
 &
   &
  GRU &
  {\color[HTML]{9B9B9B} 0.45/0.68/0.78} &
  {\color[HTML]{9B9B9B} 0.41/0.64/0.75} &
  {\color[HTML]{9B9B9B} 0.40/0.63/0.74} &
  {\color[HTML]{9B9B9B} 0.40/0.63/0.74} &
  {\color[HTML]{9B9B9B} 0.51/0.81/0.90} &
  {\color[HTML]{9B9B9B} 0.50/0.81/0.90} &
  {\color[HTML]{9B9B9B} 0.50/0.80/0.89} &
  {\color[HTML]{9B9B9B} 0.50/0.79/0.89} &
  {\color[HTML]{9B9B9B} 2.62/0.21/0.22} &
  {\color[HTML]{9B9B9B} 2.70/0.22/0.24} &
  {\color[HTML]{9B9B9B} 2.61/0.22/0.24} &
  {\color[HTML]{9B9B9B} 2.71/0.22/0.23} \\
 &
  BERTBase &
   &
  {\color[HTML]{9B9B9B} 0.76/0.93/0.97} &
  {\color[HTML]{9B9B9B} 0.74/0.92/0.96} &
  {\color[HTML]{9B9B9B} 0.74/0.92/0.96} &
  {\color[HTML]{9B9B9B} 0.74/0.92/0.96} &
  {\color[HTML]{9B9B9B} 0.64/0.94/0.98} &
  {\color[HTML]{9B9B9B} 0.64/0.94/0.98} &
  {\color[HTML]{9B9B9B} 0.64/0.94/0.97} &
  {\color[HTML]{9B9B9B} 0.64/0.94/0.98} &
  {\color[HTML]{9B9B9B} 2.75/0.13/0.16} &
  {\color[HTML]{9B9B9B} 2.64/0.28/0.34} &
  {\color[HTML]{9B9B9B} 2.68/0.26/0.32} &
  {\color[HTML]{9B9B9B} 2.64/0.28/0.35} \\
 &
  TinyBERT &
   &
  {\color[HTML]{9B9B9B} 0.77/0.93/0.96} &
  {\color[HTML]{9B9B9B} 0.74/0.92/0.96} &
  {\color[HTML]{9B9B9B} 0.74/0.92/0.96} &
  {\color[HTML]{9B9B9B} 0.74/0.92/0.96} &
  {\color[HTML]{9B9B9B} 0.64/0.94/0.97} &
  {\color[HTML]{9B9B9B} 0.65/0.94/0.98} &
  {\color[HTML]{9B9B9B} 0.64/0.94/0.97} &
  {\color[HTML]{9B9B9B} 0.65/0.94/0.97} &
  {\color[HTML]{9B9B9B} 2.54/0.28/0.35} &
  {\color[HTML]{9B9B9B} 2.60/0.29/0.36} &
  {\color[HTML]{9B9B9B} 2.51/0.30/0.36} &
  {\color[HTML]{9B9B9B} 2.60/0.29/0.35} \\
\multirow{-6}{*}{\rotatebox{90}{Unimodal(SOTA)}} &
  MobileBERT &
   &
  {\color[HTML]{9B9B9B} 0.77/0.94/0.97} &
  {\color[HTML]{9B9B9B} 0.75/0.93/0.96} &
  {\color[HTML]{9B9B9B} 0.73/0.90/0.94} &
  {\color[HTML]{9B9B9B} 0.74/0.91/0.95} &
  {\color[HTML]{9B9B9B} 0.64/0.94/0.98} &
  {\color[HTML]{9B9B9B} 0.64/0.94/0.98} &
  {\color[HTML]{9B9B9B} 0.64/0.93/0.97} &
  {\color[HTML]{9B9B9B} 0.64/0.94/0.97} &
  {\color[HTML]{9B9B9B} 2.53/0.29/0.34} &
  {\color[HTML]{9B9B9B} 2.62/0.28/0.35} &
  {\color[HTML]{9B9B9B} 2.51/0.30/0.35} &
  {\color[HTML]{9B9B9B} 2.60/0.29/0.36} \\ \hline
 &
   &
  LSTM &
  {\color[HTML]{9B9B9B} 0.76/0.94/0.97} &
  {\color[HTML]{9B9B9B} 0.75/0.93/0.96} &
  {\color[HTML]{9B9B9B} 0.39/0.62/0.73} &
  {\color[HTML]{9B9B9B} 0.75/0.92/0.96} &
  {\color[HTML]{9B9B9B} 0.66/0.94/0.98} &
  {\color[HTML]{9B9B9B} 0.66/0.95/0.98} &
  {\color[HTML]{9B9B9B} 0.66/0.94/0.98} &
  {\color[HTML]{9B9B9B} 0.66/0.94/0.98} &
  {\color[HTML]{9B9B9B} 2.63/0.21/0.22} &
  {\color[HTML]{9B9B9B} 2.72/0.21/0.22} &
  {\color[HTML]{9B9B9B} 2.53/0.31/0.37} &
  {\color[HTML]{9B9B9B} 2.64/0.30/0.36} \\
 &
   &
  RNN &
  {\color[HTML]{9B9B9B} 0.77/0.94/0.97} &
  {\color[HTML]{9B9B9B} 0.75/0.93/0.96} &
  {\color[HTML]{9B9B9B} 0.74/0.93/0.96} &
  {\color[HTML]{9B9B9B} 0.75/0.92/0.96} &
  {\color[HTML]{9B9B9B} 0.66/0.94/0.98} &
  {\color[HTML]{9B9B9B} 0.66/0.95/0.98} &
  {\color[HTML]{9B9B9B} 0.50/0.79/0.89} &
  {\color[HTML]{9B9B9B} 0.66/0.95/0.98} &
  {\color[HTML]{9B9B9B} 2.63/0.21/0.22} &
  {\color[HTML]{9B9B9B} 2.65/0.29/0.36} &
  {\color[HTML]{9B9B9B} 2.63/0.21/0.23} &
  {\color[HTML]{9B9B9B} 2.55/0.32/0.39} \\
 &
  \multirow{-3}{*}{BERTBase} &
  GRU &
  {\color[HTML]{9B9B9B} 0.77/0.94/0.97} &
  {\color[HTML]{9B9B9B} 0.75/0.93/0.96} &
  0.75{\color[HTML]{9B9B9B}/}0.93{\color[HTML]{9B9B9B}/}0.96 &
  {\color[HTML]{9B9B9B} 0.74/0.92/0.96} &
  {\color[HTML]{9B9B9B} 0.61/0.93/0.97} &
  {\color[HTML]{9B9B9B} 0.66/0.95/0.98} &
  {\color[HTML]{9B9B9B} 0.66/0.94/0.98} &
  {\color[HTML]{9B9B9B} 0.66/0.95/0.98} &
  {\color[HTML]{9B9B9B} 2.62/0.22/0.23} &
  {\color[HTML]{9B9B9B} 2.58/0.31/0.37} &
  {\color[HTML]{9B9B9B} 2.47/0.32/0.37} &
  2.52{\color[HTML]{9B9B9B}/}0.33{\color[HTML]{9B9B9B}/}0.39 \\
 &
   &
  LSTM &
  {\color[HTML]{9B9B9B} 0.77/0.93/0.97} &
  {\color[HTML]{9B9B9B} 0.74/0.92/0.96} &
  {\color[HTML]{9B9B9B} 0.74/0.92/0.96} &
  {\color[HTML]{9B9B9B} 0.74/0.92/0.96} &
  {\color[HTML]{9B9B9B} 0.66/0.94/0.98} &
  {\color[HTML]{9B9B9B} 0.64/0.94/0.98} &
  {\color[HTML]{9B9B9B} 0.66/0.94/0.98} &
  {\color[HTML]{9B9B9B} 0.65/0.94/0.98} &
  {\color[HTML]{9B9B9B} 2.46/0.33/0.38} &
  {\color[HTML]{9B9B9B} 2.56/0.32/0.37} &
  {\color[HTML]{9B9B9B} 2.45/0.33/0.38} &
  {\color[HTML]{9B9B9B} 2.55/0.32/0.37} \\
 &
   &
  RNN &
  {\color[HTML]{9B9B9B} 0.77/0.94/0.97} &
  {\color[HTML]{9B9B9B} 0.75/0.93/0.96} &
  {\color[HTML]{9B9B9B} 0.75/0.92/0.96} &
  0.75{\color[HTML]{9B9B9B}/}0.93{\color[HTML]{9B9B9B}/}0.96 &
  0.67{\color[HTML]{9B9B9B}/}0.95{\color[HTML]{9B9B9B}/}0.98 &
  {\color[HTML]{9B9B9B} 0.66/0.95/0.98} &
  {\color[HTML]{9B9B9B} 0.66/0.94/0.98} &
  {\color[HTML]{9B9B9B} 0.66/0.95/0.98} &
  {\color[HTML]{9B9B9B} 2.47/0.32/0.37} &
  {\color[HTML]{9B9B9B} 2.54/0.33/0.38} &
  {\color[HTML]{9B9B9B} 2.45/0.33/0.38} &
  {\color[HTML]{9B9B9B} 2.53/0.33/0.38} \\
 &
  \multirow{-3}{*}{TinyBERT} &
  GRU &
  {\color[HTML]{9B9B9B} 0.77/0.94/0.97} &
  {\color[HTML]{9B9B9B} 0.75/0.93/0.96} &
  {\color[HTML]{9B9B9B} 0.75/0.93/0.96} &
  {\color[HTML]{9B9B9B} 0.75/0.92/0.96} &
  {\color[HTML]{9B9B9B} 0.67/0.95/0.98} &
  0.67{\color[HTML]{9B9B9B}/}0.95{\color[HTML]{9B9B9B}/}0.98 &
  0.66{\color[HTML]{9B9B9B}/}0.94{\color[HTML]{9B9B9B}/}0.98 &
  0.67{\color[HTML]{9B9B9B}/}0.95{\color[HTML]{9B9B9B}/}0.98 &
  2.46{\color[HTML]{9B9B9B}/}0.33{\color[HTML]{9B9B9B}/}0.38 &
  {\color[HTML]{9B9B9B} 2.55/0.32/0.38} &
  {\color[HTML]{9B9B9B} 2.45/0.33/}0.39 &
  {\color[HTML]{9B9B9B} 2.53/0.33/0.38} \\
 &
   &
  LSTM &
  {\color[HTML]{9B9B9B} 0.77/0.94/0.97} &
  {\color[HTML]{9B9B9B} 0.75/0.93/0.96} &
  {\color[HTML]{9B9B9B} 0.74/0.92/0.95} &
  {\color[HTML]{9B9B9B} 0.73/0.91/0.95} &
  {\color[HTML]{9B9B9B} 0.65/0.94/0.98} &
  {\color[HTML]{9B9B9B} 0.65/0.94/0.98} &
  {\color[HTML]{9B9B9B} 0.66/0.94/0.98} &
  {\color[HTML]{9B9B9B} 0.65/0.94/0.97} &
  {\color[HTML]{9B9B9B} 2.47/0.32/0.37} &
  {\color[HTML]{9B9B9B} 2.59/0.30/0.36} &
  {\color[HTML]{9B9B9B} 2.44/0.33/0.38} &
  {\color[HTML]{9B9B9B} 2.56/0.32/0.37} \\
 &
   &
  RNN &
  {\color[HTML]{9B9B9B} 0.77/0.94/0.97} &
  {\color[HTML]{9B9B9B} 0.75/0.93/0.97} &
  {\color[HTML]{9B9B9B} 0.74/0.92/0.96} &
  {\color[HTML]{9B9B9B} 0.73/0.91/0.95} &
  {\color[HTML]{9B9B9B} 0.66/0.94/0.98} &
  {\color[HTML]{9B9B9B} 0.66/0.95/0.98} &
  {\color[HTML]{9B9B9B} 0.66/0.94/0.98} &
  {\color[HTML]{9B9B9B} 0.64/0.94/0.97} &
  {\color[HTML]{9B9B9B} 2.46/0.32/0.37} &
  2.53{\color[HTML]{9B9B9B}/}0.33{\color[HTML]{9B9B9B}/}0.38 &
  {\color[HTML]{9B9B9B} 2.45/0.33/0.38} &
  {\color[HTML]{9B9B9B} 2.55/0.32/0.37} \\
\multirow{-9}{*}{\textbf{\rotatebox{90}{Multimodal(our)}}} &
  \multirow{-3}{*}{MobileBERT} &
  GRU &
  0.77{\color[HTML]{9B9B9B}/}0.94{\color[HTML]{9B9B9B}/}0.97 &
  0.75{\color[HTML]{9B9B9B}/}0.93{\color[HTML]{9B9B9B}/}0.97 &
  {\color[HTML]{9B9B9B} 0.74/0.92/0.96} &
  {\color[HTML]{9B9B9B} 0.74/0.92/0.96} &
  {\color[HTML]{9B9B9B} 0.66/0.94/0.98} &
  {\color[HTML]{9B9B9B} 0.66/0.95/0.98} &
  {\color[HTML]{9B9B9B} 0.66/0.94/0.98} &
  {\color[HTML]{9B9B9B} 0.66/0.94/0.98} &
  {\color[HTML]{9B9B9B} 2.46/0.32/0.37} &
  {\color[HTML]{9B9B9B} 2.55/0.32/0.37} &
  2.44{\color[HTML]{9B9B9B}/}0.33{\color[HTML]{9B9B9B}/0.38} &
  {\color[HTML]{9B9B9B} 2.54/0.33/0.38} \\ \hline
\end{tabular}
}
\end{table*}

\begin{table*}[]
\centering
\caption{Compare the 3-modal EMSNet w/ and w/o progressive modality integration in tasks 1-3 on the dataset 3-modal D2 (text, vitals, image).}
\label{tab:multimodal_unimodal_d2_reduced_col_row}
\scalebox{0.7}{
\begin{tabular}{lccccccccc}
\hline
\multirow{2}{*}{}                    & \multicolumn{3}{c}{Backbone(Text/Vitals/Image)} & \multicolumn{2}{c}{Task1-Protocol (top-1/3/5)} & \multicolumn{2}{c}{Task2-MedType (top-1/3/5)} & \multicolumn{2}{c}{Task3-MedQuant (mse/pearsonr/spearmanr)} \\
                                     & T                              & V       & I    & P                      & P-M-Q                 & M                     & P-M-Q                 & Q                             & P-M-Q                       \\ \hline
\multirow{9}{*}{Fine-tuning w/o PMI} & \multirow{3}{*}{BERTBase}      & LSTM    & FC   & {\color[HTML]{9B9B9B}0.69/0.89/0.93}         & \color[HTML]{9B9B9B}0.67/0.86/0.89        & \color[HTML]{9B9B9B}0.66/0.93/0.98        & \color[HTML]{9B9B9B}0.69/0.95/0.97        & \color[HTML]{9B9B9B}1.94/0.04/-0.08               & \color[HTML]{9B9B9B}2.42/0.11/0.18              \\
                                     &                                & RNN     & FC   & \color[HTML]{9B9B9B}0.67/0.86/0.91         & \color[HTML]{9B9B9B}0.67/0.87/0.90        & \color[HTML]{9B9B9B}0.67/0.94/0.98        & \color[HTML]{9B9B9B}0.67/0.93/0.98        & \color[HTML]{9B9B9B}1.90/0.15/0.21                & \color[HTML]{9B9B9B}2.50/0.13/0.16              \\
                                     &                                & GRU     & FC   & \color[HTML]{9B9B9B}0.69/0.89/0.93         & \color[HTML]{9B9B9B}0.65/0.86/0.89        & \color[HTML]{9B9B9B}0.62/0.93/0.96        & \color[HTML]{9B9B9B}0.67/0.94/0.97        & \color[HTML]{9B9B9B}1.92/0.09/0.09                & {\color[HTML]{9B9B9B}2.15/0.16/}0.24              \\ \cline{2-10} 
                                     & \multirow{3}{*}{TinyBERT}      & LSTM    & FC   & \color[HTML]{9B9B9B}0.65/0.82/0.89         & \color[HTML]{9B9B9B}0.54/0.76/0.84        & \color[HTML]{9B9B9B}0.66/0.93/0.97        & \color[HTML]{9B9B9B}0.67/0.94/0.96        & \color[HTML]{9B9B9B}1.90/0.16/0.17                & \color[HTML]{9B9B9B}1.95/0.19/0.21              \\
                                     &                                & RNN     & FC   & \color[HTML]{9B9B9B}0.63/0.83/0.89         & \color[HTML]{9B9B9B}0.53/0.76/0.82        & \color[HTML]{9B9B9B}0.65/0.95/0.97        & \color[HTML]{9B9B9B}0.65/0.92/0.96        & 1.87\color[HTML]{9B9B9B}/0.18/0.18            & 1.93\color[HTML]{9B9B9B}/0.19/0.20              \\
                                     &                                & GRU     & FC   & \color[HTML]{9B9B9B}0.63/0.82/0.88         & \color[HTML]{9B9B9B}0.54/0.74/0.81        & \color[HTML]{9B9B9B}0.64/0.94/0.97        & \color[HTML]{9B9B9B}0.67/0.93/0.97        & \color[HTML]{9B9B9B}1.94/0.14/0.16                & {\color[HTML]{9B9B9B}1.93/}0.20{\color[HTML]{9B9B9B}/0.21}              \\ \cline{2-10} 
                                     & \multirow{3}{*}{MobileBERT}    & LSTM    & FC   & \color[HTML]{9B9B9B}0.49/0.72/0.84         & \color[HTML]{9B9B9B}0.02/0.05/0.09        & \color[HTML]{9B9B9B}0.56/0.91/0.94        & \color[HTML]{9B9B9B}0.03/0.07/0.45        & \color[HTML]{9B9B9B}4e4/-0.02/0.03                & \color[HTML]{9B9B9B}1e3/0.00/-0.02              \\
                                     &                                & RNN     & FC   & \color[HTML]{9B9B9B}0.47/0.68/0.78         & \color[HTML]{9B9B9B}0.43/0.59/0.64        & \color[HTML]{9B9B9B}0.43/0.91/0.94        & \color[HTML]{9B9B9B}0.45/0.54/0.68        & \color[HTML]{9B9B9B}1e1/0.00/-0.03                & \color[HTML]{9B9B9B}2.15/-0.03/-0.06            \\
                                     &                                & GRU     & FC   & \color[HTML]{9B9B9B}0.48/0.69/0.78         & \color[HTML]{9B9B9B}0.00/0.22/0.23        & \color[HTML]{9B9B9B}0.52/0.89/0.93        & \color[HTML]{9B9B9B}0.02/0.03/.048        & \color[HTML]{9B9B9B}3e4/0.01/-0.05                & \color[HTML]{9B9B9B}5e5/0.03/0.01               \\ \hline
\multirow{9}{*}{\textbf{Fine-tuning w/ PMI (our)}}  & \multirow{3}{*}{BERTBase}      & LSTM    & FC   & {\color[HTML]{9B9B9B}{0.72/}}0.91{\color[HTML]{9B9B9B}{/0.95}}         & {\color[HTML]{9B9B9B}0.69/0.88/}0.93        & \color[HTML]{9B9B9B}0.68/0.95/0.98        & \color[HTML]{9B9B9B}0.68/0.93/0.97        & \color[HTML]{9B9B9B}1.91/0.12/0.11                & \color[HTML]{9B9B9B}2.41/0.16/0.17              \\
                                     &                                & RNN     & FC   & \color[HTML]{9B9B9B}0.72/0.90/0.94         & \color[HTML]{9B9B9B}0.67/0.88/0.92        & \color[HTML]{9B9B9B}0.66/0.95/0.98        & 0.69\color[HTML]{9B9B9B}/0.94/0.97        & \color[HTML]{9B9B9B}1.91/0.12/0.11                & \color[HTML]{9B9B9B}2.56/0.14/0.16              \\
                                     &                                & GRU     & FC   & 0.72{\color[HTML]{9B9B9B}/0.89/}0.95         & {\color[HTML]{9B9B9B}0.68/}0.88{\color[HTML]{9B9B9B}/0.92}        & {\color[HTML]{9B9B9B}0.67/}0.96{\color[HTML]{9B9B9B}/}0.98        & {\color[HTML]{9B9B9B}0.67/}0.95{\color[HTML]{9B9B9B}/}0.98        & \color[HTML]{9B9B9B}1.91/0.11/0.10                & \color[HTML]{9B9B9B}2.30/0.16/0.21              \\ \cline{2-10} 
                                     & \multirow{3}{*}{TinyBERT}      & LSTM    & FC   & \color[HTML]{9B9B9B}0.67/0.83/0.89         & \color[HTML]{9B9B9B}0.61/0.80/0.83        & \color[HTML]{9B9B9B}0.66/0.95/0.98        & \color[HTML]{9B9B9B}0.67/0.93/0.97        & {\color[HTML]{9B9B9B}1.89/}0.18{\color[HTML]{9B9B9B}/}0.22                & \color[HTML]{9B9B9B}1.97/0.19/0.21              \\
                                     &                                & RNN     & FC   & \color[HTML]{9B9B9B}0.68/0.85/0.90         & \color[HTML]{9B9B9B}0.59/0.79/0.84        & \color[HTML]{9B9B9B}0.69/0.95/0.97        & \color[HTML]{9B9B9B}0.66/0.95/0.97        & \color[HTML]{9B9B9B}1.89/0.17/0.19                & \color[HTML]{9B9B9B}1.96/0.19/0.23              \\
                                     &                                & GRU     & FC   & \color[HTML]{9B9B9B}0.67/0.83/0.90         & \color[HTML]{9B9B9B}0.56/0.78/0.86        & 0.70\color[HTML]{9B9B9B}/0.96/0.98        & \color[HTML]{9B9B9B}0.67/0.95/0.97        & \color[HTML]{9B9B9B}1.96/0.14/0.18                & \color[HTML]{9B9B9B}1.99/0.19/0.23              \\ \cline{2-10} 
                                     & \multirow{3}{*}{MobileBERT}    & LSTM    & FC   & \color[HTML]{9B9B9B}0.72/0.89/0.94         & \color[HTML]{9B9B9B}0.64/0.86/0.89        & \color[HTML]{9B9B9B}0.66/0.95/0.98        & \color[HTML]{9B9B9B}0.68/0.93/0.97        & \color[HTML]{9B9B9B}1.93/0.13/0.14                & \color[HTML]{9B9B9B}2.12/0.15/0.18              \\
                                     &                                & RNN     & FC   & \color[HTML]{9B9B9B}0.68/0.90/0.95         & \color[HTML]{9B9B9B}0.66/0.89/0.91        & \color[HTML]{9B9B9B}0.70/0.95/0.98        & \color[HTML]{9B9B9B}0.67/0.93/0.96        & \color[HTML]{9B9B9B}1.93/0.15/0.18                & \color[HTML]{9B9B9B}2.19/0.15/0.18              \\
                                     &                                & GRU     & FC   & \color[HTML]{9B9B9B}0.70/0.88/0.93         & 0.70\color[HTML]{9B9B9B}/0.88/0.91        & \color[HTML]{9B9B9B}0.68/0.95/0.98        & \color[HTML]{9B9B9B}0.69/0.94/0.98        & \color[HTML]{9B9B9B}1.96/0.13/0.17                & \color[HTML]{9B9B9B}2.28/0.18/0.23              \\ \hline
\end{tabular}
}
\end{table*}

\subsection{Evaluation of EMSNet}

\subsubsection{2-modal EMSNet backbone accuracy on D1}

\label{sec:two_modal_backbone_evaluate}

In Table~\ref{tab:multimodal_unimodal_d1}, we compare the accuracy of the multimodal backbone with SOTA unimodal options in accomplishing tasks 1-3 on the 2-modal dataset D1. The P, M, and Q mean the backbone is trained to separately accomplish a single Task 1, 2, and 3. P-M means the backbone is trained to accomplish two tasks 1 and 2 simultaneously while P-M-Q means accomplishing three tasks at the same time. The lower mse and higher pearsonr/spearsonr indicate better performance in task3. As demonstrated, our multimodal backbone consistently outperforms unimodal options adopted in SOTA assistant systems. Specifically, it improves the top-1/3/5 accuracy for task2 by 2-3\%, reduces the MSE by $\sim$0.1, and enhances Pearsonr and Spearmanr for Task 3, which are newly proposed EMS tasks in this work.

\subsubsection{3-modal EMSNet backbone accuracy on D2}
\label{sec:three_modal_backbone_evaluate}

As discussed in Section~\ref{sec:backbone_headers}, D2(3-modal: text, vitals, scene) is a 3-modal dataset whose size (3005 samples) is about 2 orders of magnitude smaller than the dataset D1(2-modal: text, vitals). To achieve effective digestion of small-sized D2 in EMSNet's backbone, we use the progressive modality integration (PMI) to fine-tune the 3-modal backbone module on D2. Table~\ref{tab:multimodal_unimodal_d2_reduced_col_row} shows the 3-modal backbone's accuracy on D2. It's good to note that, in Table~\ref{tab:multimodal_unimodal_d2_reduced_col_row}, we only highlight 3-modal EMSNet's accuracy with and without PMI on three single tasks (P, M, Q) and three simultaneous tasks (P-M-Q), full details for two simultaneous tasks (P-M, P-Q, M-Q) can be found in Table~\ref{tab:multimodal_unimodal_d2} in Appendix~\ref{app:full_three_modal_evaluation}.

Overall, with PMI-enabled fine-tuning, 3-modal EMSNet achieves the consistently higher accuracy in recommending protocols (task1) and medicine types (task2). For example, the PMI-enabled BERTBase-LSTM-FC and BERTBase-GRU-FC achieve top-3 accuracy of 0.91 and 0.96, respectively, on the single protocol (P) and medicine type (M) tasks. For the medicine quantity recommendation task (Q and P-M-Q), although fine-tuning w/o PMI seems more performant, fine-tuning w/ PMI achieves comparable performance. For example, fine-tuning with PMI enables TinyBERT-LSTM-FC to achieve the MSE of 1.89 on the single Q task, close to the lowest MSE of 1.87. This also implies a future work direction on how PMI could further improve the regression accuracy task, like medicine quantity prescription in this paper.  

Notably, smaller 3-modal backbones that have TinyBERT or MobileBERT as the text modules are critical for on-device deployment in disaster emergency scenarios where internet access to cloud resources is destroyed. However, without PMI, these smaller models struggle with achieving high accuracy. For example, the TinyBERT-GRU-FC's top-1 accuracy is as low as 0.64 while its PMI-enabled version reaches the highest 0.7. Furthermore, given the small 3-modal dataset size in D2, the MobileBERT-GRU-FC generates random numbers in its top-3 protocol recommendation outputs, resulting in zero accuracy for three simultaneous tasks (P-M-Q). In contrast, PMI helps MobileBERT-GRU-FC's accuracy reach 0.7, which is the highest for the same metric. This indicates PMI's potential in increasing small models' multimodal capability, especially when the dataset size of different modalities is highly imbalanced (123,803 samples in D1, 3005 in D2). As one of the future works, this potential can be further explored and generalized when smaller multimodal on-device models are needed while the size of data modalities is imbalanced.









\subsubsection{Accuracy of Speech Recognition}

\begin{figure*} 
    \centering
    \begin{subfigure}{\linewidth} 
        \centering
        \includegraphics[width=\linewidth]{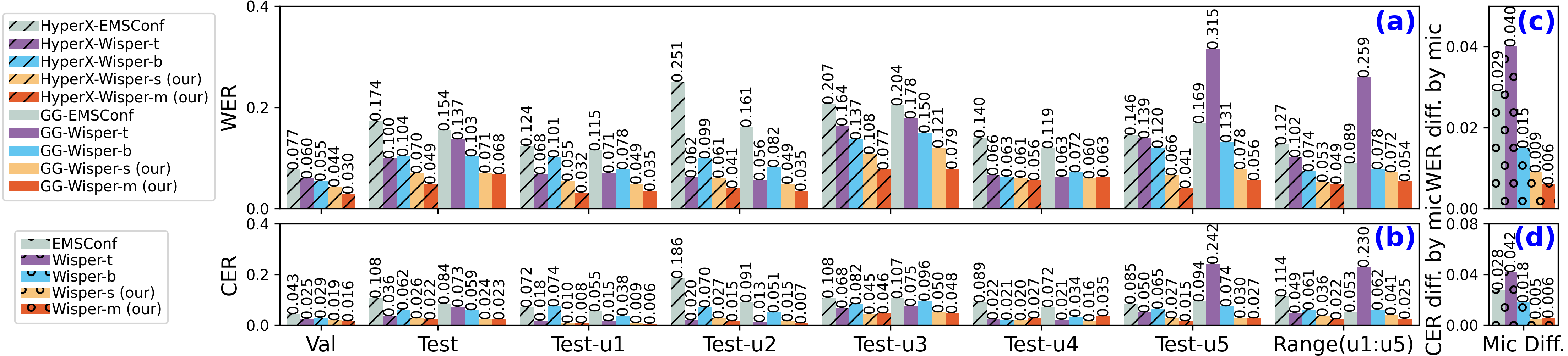}
    \end{subfigure}%
    \vspace{-10pt}
    \caption{Speech recognition evaluation of five models on both validate and test audio set.}
    \vspace{-10pt}
    \label{fig:sr_wer_cer}
\end{figure*}

Figure~\ref{fig:sr_wer_cer} demonstrates EMSWhisper's superior transcription accuracy and robustness. Our Whisper-s and Whisper-m achieve significantly lower WER and CER than three smaller SOTA speech-to-text models across validation and test sets. For instance, on user5's Google Glass (GG) test set, Whisper-m attains a WER(CER) of 0.056(0.027), whereas Whisper-t's WER(CER) rises to 0.315(0.242), about 5.6(9) times higher. As a WER of 0.1 is generally regarded as the error standard for usable voice assistants~\cite{MSTestAccMSLearn2024}, SOTA models struggle with data distribution shifts, with the test set WER exceeding 0.1 due to varying accents and microphone hardware. This challenge is even more critical in EMS settings, where stricter WER requirements render SOTA models impractical for EMS scenarios, further accrediting EMSWhisper’s ultra-low error rates. Moreover, EMSWhisper, including the Whisper-s and Whisper-m motivated by the scaling law, exhibits exceptional generalization with lower WER variance across users (Figure~\ref{fig:sr_wer_cer}(a)-(b)) and microphones(Figure~\ref{fig:sr_wer_cer}(c)-(d)), underscoring its robustness to real-world distribution shifts.

\subsubsection{Accuracy of Objection Detection}

\begin{figure}
    \includegraphics[width=0.95\linewidth]{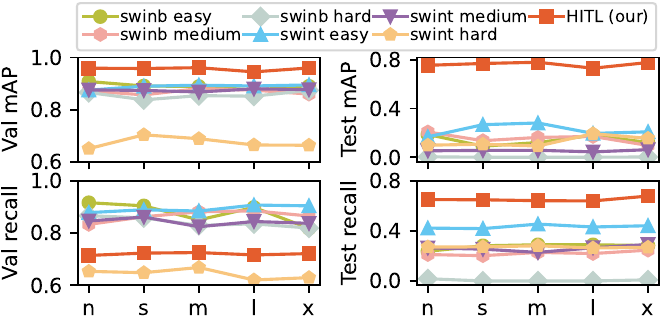}
    \vspace{-10pt}
\caption{Evaluation of fine-tuning five YOLO11 models (nano, small, medium, large, extra large) on images labeled by 2 Grounding Dino models (swinb, swint) and labeled with our human-in-the-loop (HITL) annotation adjustment.}
    \label{fig:object_detection_results}
\end{figure}

Figure~\ref{fig:object_detection_results} compares the object detection performance of fine-tuning YOLO11~\cite{YOLO11Ultralytics2024} on the image dataset D4(image) labeled by Grounding Dino and labeled by our human-in-the-loop (HITL) annotation adjustment. Our HITL annotation strategy outperforms Grounding Dino's annotations, bringing the test mAP value close to 0.8. In contrast, the mAP of all YOLO11 models on images annotated with Grounding Dino is below 0.6, which is unacceptable. The high mAP on the validate set but low mAP and recall on the test set means fine-tuning YOLO11 with Grounding Dino's annotations causes the overfitting during the fine-tuning process. For example, using Grounding Dino with the hard text prompts from Figure~\ref{fig:gd_prompts_map_recall_vertical} can achieve an mAP above 0.8, but the test mAP of all YOLO11 models is nearly 0. The recall metric is similar: high validate recall while much lower test recall. This results from Grounding Dino's high false positive predictions as discussed in Section~\ref{sec:background}. Our HITL avoids this problem because we add the human adjustments to correct Grounding Dino's false positive annotations, enabling much higher test mAP and recall. It's important to note that when compared to direct manual annotations,
the advantage of our HITL is that we exploit Grounding Dino's accurate (i.e., true positives) annotations to save human labor time on annotations.

\subsubsection{Accuracy of OCR and Barcode scanner}

\begin{figure}[h]
    \includegraphics[width=1\linewidth]{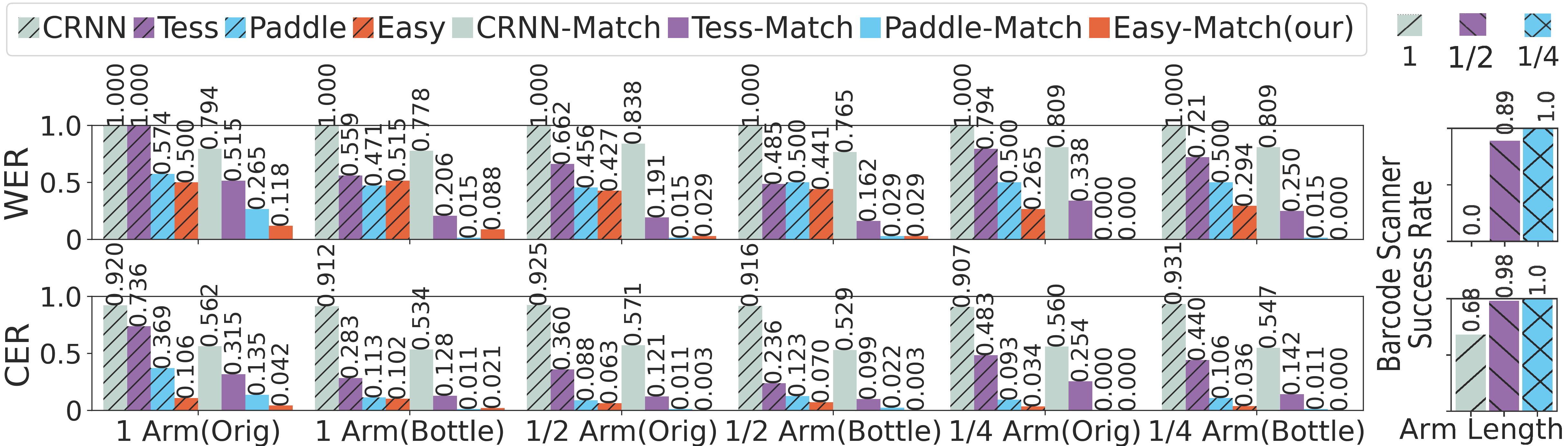}
    \vspace{-15pt}
    \caption{(left) Evaluation of four OCR models (CRNN, TesseractOCR, PaddleOCR, EasyOCR) on medicine bottle images and crops of the image (original, bottle). (right) Barcode reader success rate at different arm lengths for originals (top) and bottles (bottom).}
    \vspace{-8pt}
    \label{fig:ocr_results}
\end{figure}

For the OCR module in EMSGlass, we evaluate an image dataset of size 204 captured by Google Glass's 8MP camera, featuring a user holding a labeled medicine bottle at three distances—full arm ($\sim$0.6m), half arm ($\sim$ 0.3m), and quarter arm ($\sim$0.15m)—to account for varying EMT arm lengths. Each image has two versions: the original image without cropping and the cropped bottle segment. We use the word error rate (WER) and character error rate (CER) to measure OCR accuracy. As shown in Figure \ref{fig:ocr_results}, EMSGlass OCR performance was evaluated using four state-of-the-art (SOTA) models listed in Table~\ref{tab:emsfoundation_components}: EasyOCR, TesseractOCR, PaddleOCR, CRNN. In general, EasyOCR achieves the lowest WER and CER. Our edit distance (ED)-based matching discussed in Section~\ref{sec:ocr_barcode_medmath} is applied after each OCR prediction. As shown, the OCR models with ED-match can achieve significantly decreased WER and CER, indicating the effectiveness of our proposed ED-match method. ED-match helps EasyOCR to decrease WER  and CER by 89\% and 83\%, respectively. EMSGlass employs Easy-Match as its OCR framework, which consistently outperforms other models with WER (CER) below 0.12 (0.05) across all arm distance and cropping conditions. 

For the barcode scanner, we only use the original image without cropping. In Figure~\ref{fig:ocr_results} (right), we use the success rate as the metric to measure the performance of barcode scanner (developed based on ML Kit~\cite{MLKit} in Android). At the 1/4 Arm, our scanner can achieve a 100\% accuracy, indicating its effectiveness. On average, it takes less than 0.5 seconds to process one image, ensuring minimal latency.

\begin{table}
\centering
\caption{End2End EMSFoundation model accuracy with different speech recognition (SR) and objection detection (OD) models. We use MobileBERT, GRU, and FC in the backbone model.}
\vspace{-8pt}
\label{tab:e2e_accuracy}
\scalebox{0.51}{
\begin{tabular}{lllll}
\hline
 &
   &
  Task1: Protocol &
  Task2: Med-type &
  Task3: Med-quant \\
\multirow{-2}{*}{SR} &
  \multirow{-2}{*}{OD} &
  Top-1/3/5 &
  Top-1/3-5 &
  mse/pearsonr/spearmanr \\ \hline
\multicolumn{2}{c}{MobileBERT} &
   &
   &
   \\
Truth &
   &
  {\color[HTML]{9B9B9B} 0.70/0.90/0.94} &
  {\color[HTML]{9B9B9B} 0.54/0.94/0.96} &
  {\color[HTML]{9B9B9B} 1.67/0.29/0.23} \\
Whisper-s &
   &
  {\color[HTML]{9B9B9B} 0.70/0.92/0.98}$^{+0.00/+0.02/+0.04}$ &
  {\color[HTML]{9B9B9B} 0.55/0.95/0.98}$^{+0.01/+0.01/+0.02}$ &
  {\color[HTML]{9B9B9B} 1.66/0.28/0.29}$^{-0.01/-0.01/+0.06}$ \\
Whisper-m &
   &
  {\color[HTML]{9B9B9B} 0.70/0.90/0.95}$^{+0.00/+0.00/-0.01}$ &
  {\color[HTML]{9B9B9B} 0.56/0.95/0.98}$^{+0.02/+0.01/+0.02}$ &
  {\color[HTML]{9B9B9B} 1.66/0.28/0.27}$^{-0.01/-0.01/+0.04}$ \\ \hline
\multicolumn{2}{c}{MobileBERT-GRU} &
   &
   &
   \\
Truth &
   &
  {\color[HTML]{9B9B9B} 0.72/0.90/0.96} &
  {\color[HTML]{9B9B9B} 0.50/0.94/0.96} &
  {\color[HTML]{9B9B9B} 1.66/0.32/0.22} \\
Whisper-s &
   &
  {\color[HTML]{9B9B9B} 0.71/0.91/0.97}$^{-0.01/+0.01/+0.01}$ &
  {\color[HTML]{9B9B9B} 0.51/0.94/0.98}$^{+0.01/+0.00/+0.02}$ &
  {\color[HTML]{9B9B9B} 1.56/0.39/0.32}$^{-0.1/+0.07/+0.10}$ \\
Whisper-m &
   &
  {\color[HTML]{9B9B9B} 0.71/0.90/0.96}$^{-0.01/+0.00/+0.00}$ &
  {\color[HTML]{9B9B9B} 0.52/0.94/0.97}$^{+0.02/+0.00/+0.01}$ &
  {\color[HTML]{9B9B9B} 1.56/0.39/0.30}$^{-0.1/+0.07/+0.08}$ \\ \hline
\multicolumn{2}{c}{MobileBERT-GRU-FC} &
   &
   &
   \\
Truth &
   &
  {\color[HTML]{9B9B9B} 0.76/0.96/0.98} &
  {\color[HTML]{9B9B9B} 0.58/0.92/0.98} &
  {\color[HTML]{9B9B9B} 1.61/0.36/0.34} \\
Whisper-s &
   &
  {\color[HTML]{9B9B9B} 0.74/0.97/1.00}$^{-0.02/+0.01/+0.02}$ &
  {\color[HTML]{9B9B9B} 0.61/0.92/0.99}$^{+0.03/+0.00/+0.01}$ &
  {\color[HTML]{9B9B9B} 1.60/0.36/0.33}$^{-0.01/+0.00/-0.01}$ \\
Whisper-m &
  \multirow{-3}{*}{Truth} &
  {\color[HTML]{9B9B9B} 0.73/0.98/1.00}$^{-0.03/+0.02/+0.02}$ &
  {\color[HTML]{9B9B9B} 0.61/0.92/0.99}$^{+0.03/+0.00/+0.01}$ &
  {\color[HTML]{9B9B9B} 1.59/0.36/0.32}$^{-0.02/+0.00/-0.02}$ \\ \hline
\multicolumn{2}{c}{MobileBERT-GRU-FC} &
   &
   &
   \\
Truth &
   &
  {\color[HTML]{9B9B9B} 0.76/0.96/0.98} &
  {\color[HTML]{9B9B9B} 0.58/0.92/0.98} &
  {\color[HTML]{9B9B9B} 1.61/0.36/0.34} \\
Whisper-s &
   &
  {\color[HTML]{9B9B9B} 0.74/0.97/1.00}$^{-0.02/+0.01/+0.02}$ &
  {\color[HTML]{9B9B9B} 0.61/0.92/0.99}$^{+0.03/+0.00/+0.01}$ &
  {\color[HTML]{9B9B9B} 1.60/0.36/0.33}$^{-0.01/+0.00/-0.01}$ \\
Whisper-m &
  \multirow{-3}{*}{yolo11n} &
  {\color[HTML]{9B9B9B} 0.73/0.98/1.00}$^{-0.03/+0.02/+0.02}$ &
  {\color[HTML]{9B9B9B} 0.61/0.92/0.99}$^{+0.03/+0.00/+0.01}$ &
  {\color[HTML]{9B9B9B} 1.59/0.36/0.32}$^{-0.02/+0.00/-0.02}$ \\ \hline
\multicolumn{2}{c}{MobileBERT-GRU-FC} &
   &
   &
   \\
Truth &
   &
  {\color[HTML]{9B9B9B} 0.76/0.96/0.98} &
  {\color[HTML]{9B9B9B} 0.58/0.92/0.98} &
  {\color[HTML]{9B9B9B} 1.61/0.36/0.34} \\
Whisper-s &
   &
  {\color[HTML]{9B9B9B} 0.74/0.97/1.00}$^{-0.02/+0.01/+0.02}$ &
  {\color[HTML]{9B9B9B} 0.61/0.92/0.99}$^{+0.03/+0.00/+0.01}$ &
  {\color[HTML]{9B9B9B} 1.60/0.36/0.33}$^{-0.01/+0.00/-0.01}$ \\
Whisper-m &
  \multirow{-3}{*}{yolo11x} &
  {\color[HTML]{9B9B9B} 0.73/0.98/1.00}$^{-0.03/+0.02/+0.02}$ &
  {\color[HTML]{9B9B9B} 0.61/0.92/0.99}$^{+0.03/+0.00/+0.01}$ &
  {\color[HTML]{9B9B9B} 1.59/0.36/0.32}$^{-0.02/+0.00/-0.02}$ \\ \hline
\end{tabular}
}
\end{table}

\subsubsection{End-to-End accuracy on D2(3-modal,text,vitals,image)} 

Table~\ref{tab:e2e_accuracy} shows EMSNet's end-to-end (E2E) accuracy on the 3-modal dataset D2, i.e., from the end of original 3-modal inputs to the end of tasks 1-3. Here we use our Whisper-s and Whisper-m for the speech recognition (SR) module, YOLO11n for the object detection (OD) module, and 3 models in the backbone: unimodal MobileBERT, 2-modal MobileBERT-GRU, and 3-modal MobileBERT-GRU-FC. We make two observations from the table: 1) the multimodal models, including the 2-modal and 3-modal models, achieve higher accuracy, 2) the addition of speech recognition and object detection models don't degrade the E2E accuracy, indicating seamless integrations with the backbone models.

\subsection{Evaluation of EMSServe}
\subsubsection{Evaluation plan} EMSServe is the first multimodal model serving framework addressing the issue of asynchronous arrival times, which is inherently introduced by using smart glasses in multimodal EMS scenarios with high mobility. Here, we make the following evaluation plan to show EMSServe's advantage in serving such asynchronously arrived multimodal data. Note that, when evaluating EMSServe, the BERTBase-GRU-FC backbone is used with Whisper-tiny and YOLO11n as the multimodal EMSNet model.

\textbf{Three episodes}: As shown in Table~\ref{tab:thee_episode}, we evaluate EMSServe by running three episodes of multimodal data arriving at different timestamps. Episode 1 includes one speech data, followed by ten continuous vitals data, and then ends up with ten continuous image data. Episode 1 echoes the typical data arrival sequence illustrated in the aforementioned Figure~\ref{fig:different_time_arrival} and Figure~\ref{fig:emsserve_oview}. Episodes 2 and 3 randomly shuffle the data sequence in episode 1 with two different random seeds. These three episodes collectively cover different scene data arrival times of multimodal data in real-world EMS events.

\begin{table}[h]
\centering
\caption{Three episodes contain three asynchronously arrived multimodal data sequences, reflecting different arrival times of real-world EMS scene data.}
\label{tab:thee_episode}
\scalebox{0.62}{
\begin{tabular}{|c|ccccccccccccccccccccc|}
\hline
Episode &
  \multicolumn{21}{c|}{Asynchronously arrived multimodal data sequence (S-Speech, V-Vital, I-Image)} \\ \hline
1 &
  \multicolumn{1}{c|}{\cellcolor[HTML]{FFCCC9}S} &
  \multicolumn{1}{c|}{\cellcolor[HTML]{ECF4FF}V} &
  \multicolumn{1}{c|}{\cellcolor[HTML]{ECF4FF}V} &
  \multicolumn{1}{c|}{\cellcolor[HTML]{ECF4FF}V} &
  \multicolumn{1}{c|}{\cellcolor[HTML]{ECF4FF}V} &
  \multicolumn{1}{c|}{\cellcolor[HTML]{ECF4FF}V} &
  \multicolumn{1}{c|}{\cellcolor[HTML]{ECF4FF}V} &
  \multicolumn{1}{c|}{\cellcolor[HTML]{ECF4FF}V} &
  \multicolumn{1}{c|}{\cellcolor[HTML]{ECF4FF}V} &
  \multicolumn{1}{c|}{\cellcolor[HTML]{ECF4FF}V} &
  \multicolumn{1}{c|}{\cellcolor[HTML]{ECF4FF}V} &
  \multicolumn{1}{c|}{\cellcolor[HTML]{D9D2E9}I} &
  \multicolumn{1}{c|}{\cellcolor[HTML]{D9D2E9}I} &
  \multicolumn{1}{c|}{\cellcolor[HTML]{D9D2E9}I} &
  \multicolumn{1}{c|}{\cellcolor[HTML]{D9D2E9}I} &
  \multicolumn{1}{c|}{\cellcolor[HTML]{D9D2E9}I} &
  \multicolumn{1}{c|}{\cellcolor[HTML]{D9D2E9}I} &
  \multicolumn{1}{c|}{\cellcolor[HTML]{D9D2E9}I} &
  \multicolumn{1}{c|}{\cellcolor[HTML]{D9D2E9}I} &
  \multicolumn{1}{c|}{\cellcolor[HTML]{D9D2E9}I} &
  \cellcolor[HTML]{D9D2E9}I \\ \hline
2 &
  \multicolumn{1}{c|}{\cellcolor[HTML]{D9D2E9}I} &
  \multicolumn{1}{c|}{\cellcolor[HTML]{ECF4FF}V} &
  \multicolumn{1}{c|}{\cellcolor[HTML]{D9D2E9}I} &
  \multicolumn{1}{c|}{\cellcolor[HTML]{ECF4FF}V} &
  \multicolumn{1}{c|}{\cellcolor[HTML]{D9D2E9}I} &
  \multicolumn{1}{c|}{\cellcolor[HTML]{ECF4FF}V} &
  \multicolumn{1}{c|}{\cellcolor[HTML]{D9D2E9}I} &
  \multicolumn{1}{c|}{\cellcolor[HTML]{FFCCC9}S} &
  \multicolumn{1}{c|}{\cellcolor[HTML]{ECF4FF}V} &
  \multicolumn{1}{c|}{\cellcolor[HTML]{D9D2E9}I} &
  \multicolumn{1}{c|}{\cellcolor[HTML]{ECF4FF}V} &
  \multicolumn{1}{c|}{\cellcolor[HTML]{D9D2E9}I} &
  \multicolumn{1}{c|}{\cellcolor[HTML]{D9D2E9}I} &
  \multicolumn{1}{c|}{\cellcolor[HTML]{ECF4FF}V} &
  \multicolumn{1}{c|}{\cellcolor[HTML]{ECF4FF}V} &
  \multicolumn{1}{c|}{\cellcolor[HTML]{D9D2E9}I} &
  \multicolumn{1}{c|}{\cellcolor[HTML]{ECF4FF}V} &
  \multicolumn{1}{c|}{\cellcolor[HTML]{ECF4FF}V} &
  \multicolumn{1}{c|}{\cellcolor[HTML]{D9D2E9}I} &
  \multicolumn{1}{c|}{\cellcolor[HTML]{ECF4FF}V} &
  \cellcolor[HTML]{D9D2E9}I \\ \hline
3 &
  \multicolumn{1}{c|}{\cellcolor[HTML]{ECF4FF}V} &
  \multicolumn{1}{c|}{\cellcolor[HTML]{ECF4FF}V} &
  \multicolumn{1}{c|}{\cellcolor[HTML]{ECF4FF}V} &
  \multicolumn{1}{c|}{\cellcolor[HTML]{ECF4FF}V} &
  \multicolumn{1}{c|}{\cellcolor[HTML]{ECF4FF}V} &
  \multicolumn{1}{c|}{\cellcolor[HTML]{ECF4FF}V} &
  \multicolumn{1}{c|}{\cellcolor[HTML]{D9D2E9}I} &
  \multicolumn{1}{c|}{\cellcolor[HTML]{D9D2E9}I} &
  \multicolumn{1}{c|}{\cellcolor[HTML]{D9D2E9}I} &
  \multicolumn{1}{c|}{\cellcolor[HTML]{D9D2E9}I} &
  \multicolumn{1}{c|}{\cellcolor[HTML]{D9D2E9}I} &
  \multicolumn{1}{c|}{\cellcolor[HTML]{D9D2E9}I} &
  \multicolumn{1}{c|}{\cellcolor[HTML]{ECF4FF}V} &
  \multicolumn{1}{c|}{\cellcolor[HTML]{D9D2E9}I} &
  \multicolumn{1}{c|}{\cellcolor[HTML]{ECF4FF}V} &
  \multicolumn{1}{c|}{\cellcolor[HTML]{ECF4FF}V} &
  \multicolumn{1}{c|}{\cellcolor[HTML]{D9D2E9}I} &
  \multicolumn{1}{c|}{\cellcolor[HTML]{D9D2E9}I} &
  \multicolumn{1}{c|}{\cellcolor[HTML]{FFCCC9}S} &
  \multicolumn{1}{c|}{\cellcolor[HTML]{ECF4FF}V} &
  \cellcolor[HTML]{D9D2E9}I \\ \hline
\end{tabular}
}
\end{table}

\textbf{Three scenarios}: We run episodes in the following three EMS scenarios and measure the cumulative time spent in inference. To avoid measurement interferences like cold-start, we perform all experiments 15 times, and take the average results from the last 10 runs. Scenarios 2) and 3) below employ the feature cache in EMSServe by default.

\begin{figure*}[h]
    \centering
    \includegraphics[width=\textwidth]{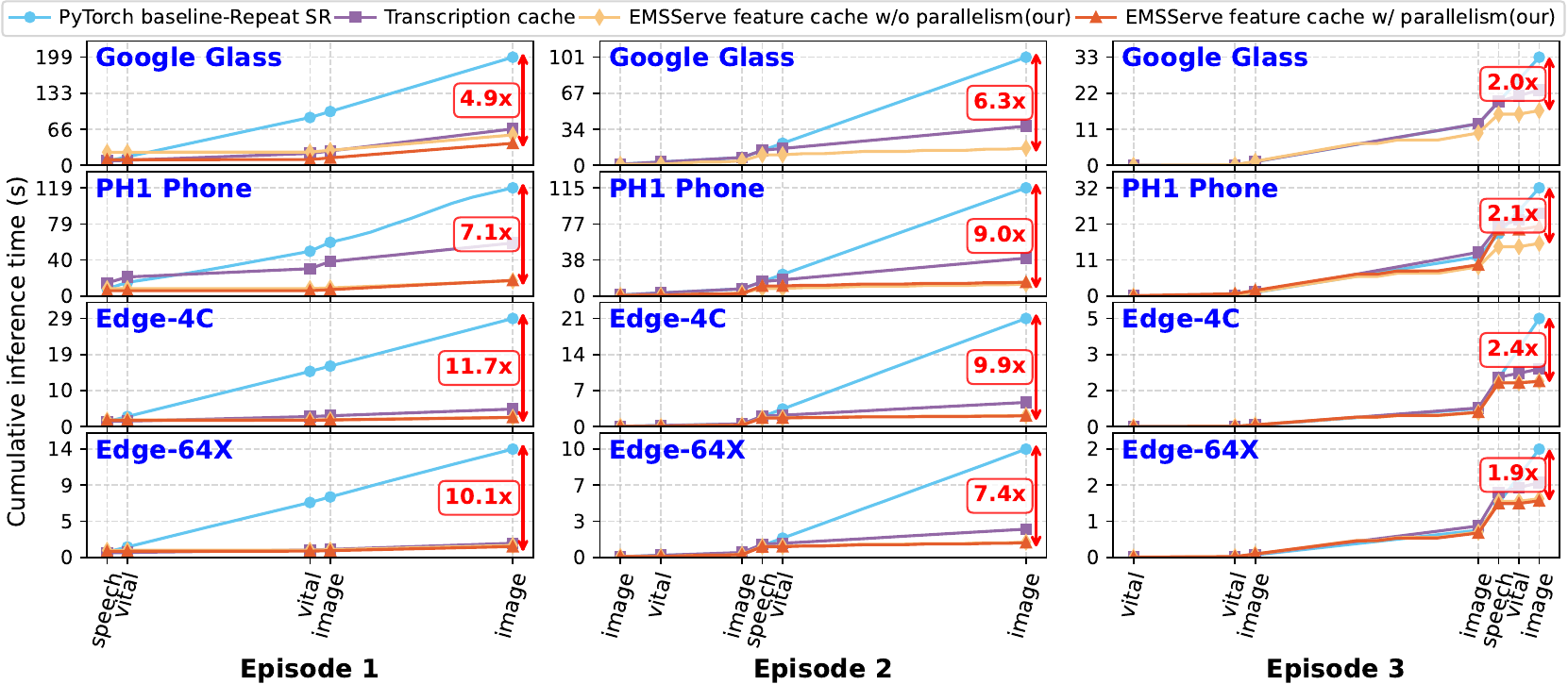}
    \vspace{-20pt}
    \caption{In scenario 1, our EMSServe achieves the lowest cumulative inference time across the episode on all hardware platforms: (a)Google Glass, (b)PH1 Phone, (c)Edge-4C, (d)Edge-64x.}
    \vspace{-8pt}
    \label{fig:static_e2e_10vitals_10scenes}
\end{figure*}

\begin{figure*}[h]
    \centering
    \includegraphics[width=\textwidth]{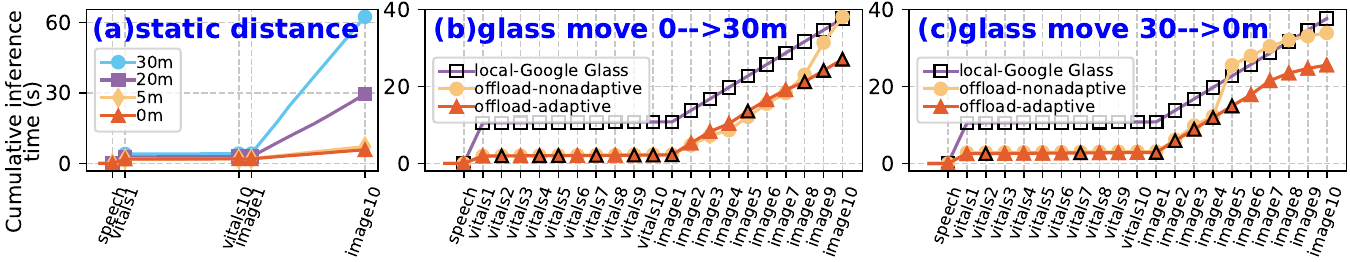}
    \vspace{-20pt}
    \caption{Evaluation of offloading design w/o mobility (a) and w mobility (b-c).}
    \vspace{-8pt}
    \label{fig:nonadapative_adaptive_offloading}
\end{figure*}

\textbf{Scenario \#1: static running without offloading}. In Figure~\ref{fig:static_e2e_10vitals_10scenes}, we statically run three episodes in Table~\ref{tab:thee_episode} on four real-world hardware specified in Table~\ref{tab:hardware}, including Google Glass, PH1 Phone, and 2 edge servers: Edge-4C and Edge-64X. Directly using PyTorch for multimodal inference on mobile devices and edge servers linearly increases cumulative inference time costs. In contrast, our EMSServe demonstrates much better scalability, incurring minimal cumulative inference time increments in all episodes. This is because the conventional multimodal serving framework repeats text module computation across a whole episode. With the modality-aware splitter splitting the text module and other modal modules apart, EMSServe can cache the text features after processing early arrived symptom speech, and reuse the text feature cache with lately arrived data for subsequent multimodal inference. This avoids repeated text feature cache computations in EMS scenarios where different data arrives asynchronously, achieving 1.9× – 11.7× speedup on all hardware platforms. In real-world deployments where the length of an episode is longer (e.g., unlimited images in video streaming), EMSServe's advantage will be more prominent. 

\textbf{Scenario \#2: offloading without user mobility}. When compared to Google Glass, Edge-4C's lower inference latency illustrated in Figure~\ref{fig:backbone_components_inference_time} and Figure~\ref{fig:static_e2e_10vitals_10scenes} expose opportunities to offload compute-expensive inference workloads for the benefit of faster inference. We use episode 1 to evaluate the impact of non-line-of-sight (NLOS) distances in inference workloads offloading from Google Glass to Edge-4C, as illustrated in Figure~\ref{fig:ui_design}. We put the Google Glass at static distances (0-30 meters) to the Edge-4C in a building. This scenario \#2 reflects a typical EMS realism, where the EMTs wearing smart glasses and the EMTs carrying the manpack collaboratively search for patients and perform EMS interventions in different indoor rooms. Figure~\ref{fig:nonadapative_adaptive_offloading}(a) provides the evaluation results, where 5m indicates an NLOS room between Google Glass and Edge-4C while 30m indicates 6 rooms. As we increase the distances, offloading the symptom speech or vitals produces similar cumulative inference latencies while offloading the 10 images tells the essence of offloading in EMS scenarios: it's more advantageous to offload images to Edge-4C when the distances are long(e.g., >5m). This is mainly because the size of images used in the episode is relatively large, making offloading images sensitive to the glass-edge distances.

\textbf{Scenario \#3: offloading with user mobility}. Following up with scenario \#2 discussed above, a user wearing Google Glass walks from 0m to 30m and then walks back from 30m to 0m, evaluating the adaptive offloading enabled by the latency monitor thread in EMSServe. This is important because, as discussed in Section~\ref{sec:intro}, seamless EMS recommendations on the UI display require lower inference latencies while the user moves in EMS scenes. The black triangles in Figures~\ref{fig:nonadapative_adaptive_offloading}(b-c) indicate the on-glass inference when adaptive offloading is employed. Both figures illustrate that decisions on offloading vitals are not sensitive to mobility due to the fast vitals inference on Google Glass. In (b), with automatic on-glass inference enabled by our adaptive offloading design, where image8 distinguishes the adaptive and nonadaptive offloading options, where the adaptive offloading provides lower model serving latency. In (c), the speech arrives at 30m. While the EMT walks towards Edge-4C, the vitals and images arrive in sequence. When the distance is relatively long (image1-image4), the offloading options achieve comparable cumulative serving latency to on-glass options. When the wireless network becomes unstable (image5), the advantage of adaptively choosing on-glass serving becomes prominent, leaving an obvious gap between non-adaptive and adaptive serving.

\subsection{User study with EMTs}

To evaluate EMSGlass's usability and demonstrate its real-world end-to-end usage feasibility, we conducted a user study with 6 certified Emergency Medical Technicians (EMTs). Each EMT uses the full EMSGlass system (i.e., EMSNet and EMSServe) in a simulated emergency room illustrated in Figure~\ref{fig:user_study_room}(e) in Appendix~\ref{sec:user_study_room}. The user study process involves data perception and real-time decision-making in a simulated end-to-end EMS event, including protocol selection and medication prescription tasks. Following the user study, participants completed a 15-item Likert scale questionnaire (1 = strongly disagree, 5 = strongly agree) covering five dimensions:

\begin{itemize}
    \item \textbf{U}: Usability and user experience
    \item \textbf{P}: Presentation of input modalities and task outputs
    \item \textbf{E}: Effectiveness of multimodal fusion
    \item \textbf{W}: Workflow integration 
    \item \textbf{F}: Improvement suggestions 
\end{itemize}

Participants also took part in semi-structured post-study interviews. The semi-structured interviews revealed rich, context-specific perspectives on user adoption, which are especially more meaningful to the HCI and mobile computing communities. In light of this, we will briefly highlight the quantitative scores while focusing more on the qualitative insights. More details regarding specific user study processes and quantitative results can be found in Appendix~\ref{sec:user_study}.

\subsubsection{Quantitative results}

Figure~\ref{fig:user_study_quant_scores} in Appendix~\ref{sec:user_study_quant_assessment} presents detailed scores. Overall, EMSGlass achieved consistently high ratings across all five evaluation dimensions, demonstrating its strong usability and technical reliability in simulated end-to-end EMS scenarios. Participants rated usability and interaction positively, with average scores between 4.0 and 4.3, indicating that the system was easy to operate during time-sensitive tasks. Multimodal perception accuracy emerged as a particular strength: voice transcription, pill and alcohol detection, and especially medicine label recognition achieved mean scores between 4.3 and 5, reflecting highly reliable system performance. The clarity and correctness of system recommendations throughout the scenario also received favorable ratings (average above 4.0), confirming that participants were able to follow and trust EMSGlass’s evolving guidance during emergency response tasks. Taken together, these quantitative results highlight EMSGlass’s potential to deliver accurate, real-time, and user-friendly decision support for emergency medical care.

\subsubsection{Qualitative insights} Overall, the qualitative findings highlight a clear pattern: EMSGlass is technically robust in multimodal recognition, but practical deployment hinges on hardware ergonomics, system responsiveness, workflow flexibility, and ecosystem integration.

\textbf{1) Clinical utility vs. Field realities.} Participants praised EMSGlass’s automated med math and protocol recommendation capabilities, describing them as ``very useful'' for quick calculations under pressure. However, they emphasized that field conditions often diverge from lab simulations:

``There’s a big gap between this controlled study and the real world---rainy days, low light, or rapidly moving a patient on a trail. The glasses could easily fall off in those scenarios.''(User1) They also noted that medication choices often stem from written protocols, not visual cues, and that misalignment here, if it exists, can potentially reduce trust.

\textbf{2) Integration with communication and documentation systems.} Multiple participants identified integration with existing EMS tools (e.g., Pulsara for documentation, Zoll for vitals, incident command dashboards) as a key driver of real-world utility and adoption:

``If it could fill out Pulsara automatically, it’d save a lot of time during handover.'' (User2) ``Streaming scene video to incident command would help coordinate resources during big events.'' (User1) These insights highlight EMSGlass’s potential to reduce communication overhead and automate cumbersome documentation, which are both critical pain points in EMS operations.

\textbf{3) Hardware ergonomics and interaction design.} Hardware limitations were a recurring theme. Participants described the current Google Glass hardware as ``outdated'', citing fit issues, instability during movement, and awkward display positioning. Suggestions included rear support bands, hardware buttons usable with gloves, and integration with newer AR glasses platforms:

``It works okay when stationary, but during emergencies it just doesn’t stay put.'' (User3) Interaction-wise, users expressed a preference for press-and-hold voice input, familiar from consumer remotes, over toggling gestures.

\textbf{4) Customization for local agencies.}
Participants highlighted that EMS protocols and medication availability vary widely by region, and EMSGlass should support local customization:

``Every area has different meds. EMTs and local doctors control what’s available — one size won’t fit all.'' (User2) This points to the need for flexible, user-editable protocol and medication support for widespread adoption.

\textbf{5) Adoption drivers: accuracy, speed, reliability.} Across interviews, participants consistently identified accuracy, real-time speed, and reliability as the three core adoption drivers. Medication recognition accuracy was viewed positively, but slow scene processing and occasional recommendation mismatches were cited as the direction for hardware and software improvements in the future:

``Vitals are crucial. If I’m moving fast, it needs to still keep up.'' (User3) ``Faster and more accurate would definitely make me more willing to use it in real cases.'' (User1)

\section{Conclusions}
\label{sec:discuss_conclusion}
We present EMSGlass, a networked smart glass system designed to optimize the EMTs' workflow by leveraging a multimodal foundation model, EMSFoundation, and an
adaptive edge-assisted multimodal model serving framework, EMSServe. EMSFoundation is trained on a real-world multimodal EMS dataset to accomplish up to five critical EMS tasks. EMSServe exploits the feature cache during serving multimodal models to handle variable multimodal data arrival times. Extensive experiments and evaluations demonstrate EMSGlass's effectiveness, showing significant improvements in both decision accuracy and processing speed.


\bibliographystyle{unsrt}
\bibliography{references}

\appendix
\section{Multimodal Dataset Processor}
\label{sec:multimodal_processor_implement}

As discussed in Section~\ref{sec:multimodal_processor_design}, our multimodal data processor includes the following steps for vitals data: 
\begin{enumerate}
    \item Outlier removal: due to unattended mistakes during recording the patient vitals in real-world EMS events, raw vitals data in NEMSIS often contain default maximum or minimum values, e.g.,  heart rate at 500 per minute. To avoid the influence of these  unrealistic extreme vitals, e.g., maximum default values, we apply the 2nd and 98th percentile clipping, removing the vitals outside the 2\%-98\% percentile range.  
    \item Padding: for the missed values in raw NEMSIS vitals, we simply add zero at the beginning of the vitals. Our experiments show this practice enables EMSNet to achive higher accuracy.
    \item Cross-sample normalization: NEMSIS dictates different scale ranges for different vitals, e.g., pulse oximetry (PO) in [0, 100] while blood glucose in [0, 2000]. These large numerical values produces the notorious ``NaN'' problems during training~\cite{google_normalization}. In addition, different scales of vitals values prevent deep learning models to effectively combine the information from different vitals. To address this problem, we adopt three common normalization methods: z-score,min-max and min-max over z-score~\cite{google_normalization,rahmad2024comparativenorm}.
\end{enumerate}

We follow the design in EMSAssist~\cite{EMSAssist2023MobiSys} to set the number of protocols at 46. We take all the samples and set the number of medicine at 18. We apply a one-hot encoding process to both protocol and medicine type labels. We apply the same outlier removal and normalizations steps of vitals to the medicine quantity value labels. As medicine quantity labels are single float numbers, we did not apply the padding step, which is only required by the time-series vitals. Thus, task1 and task2 are both essentially single-label multi-class classification problems while the task3 is a regression problem. In the end, we obtain 123,803 samples as D1 that contain 2-modes of inputs (text and vitals) and corresponding labels for three tasks. We further split it into train/validate/test sets with 74821/24761/24761 (3:1:1) samples, respectively.

\begin{figure}
    \includegraphics[width=\linewidth]{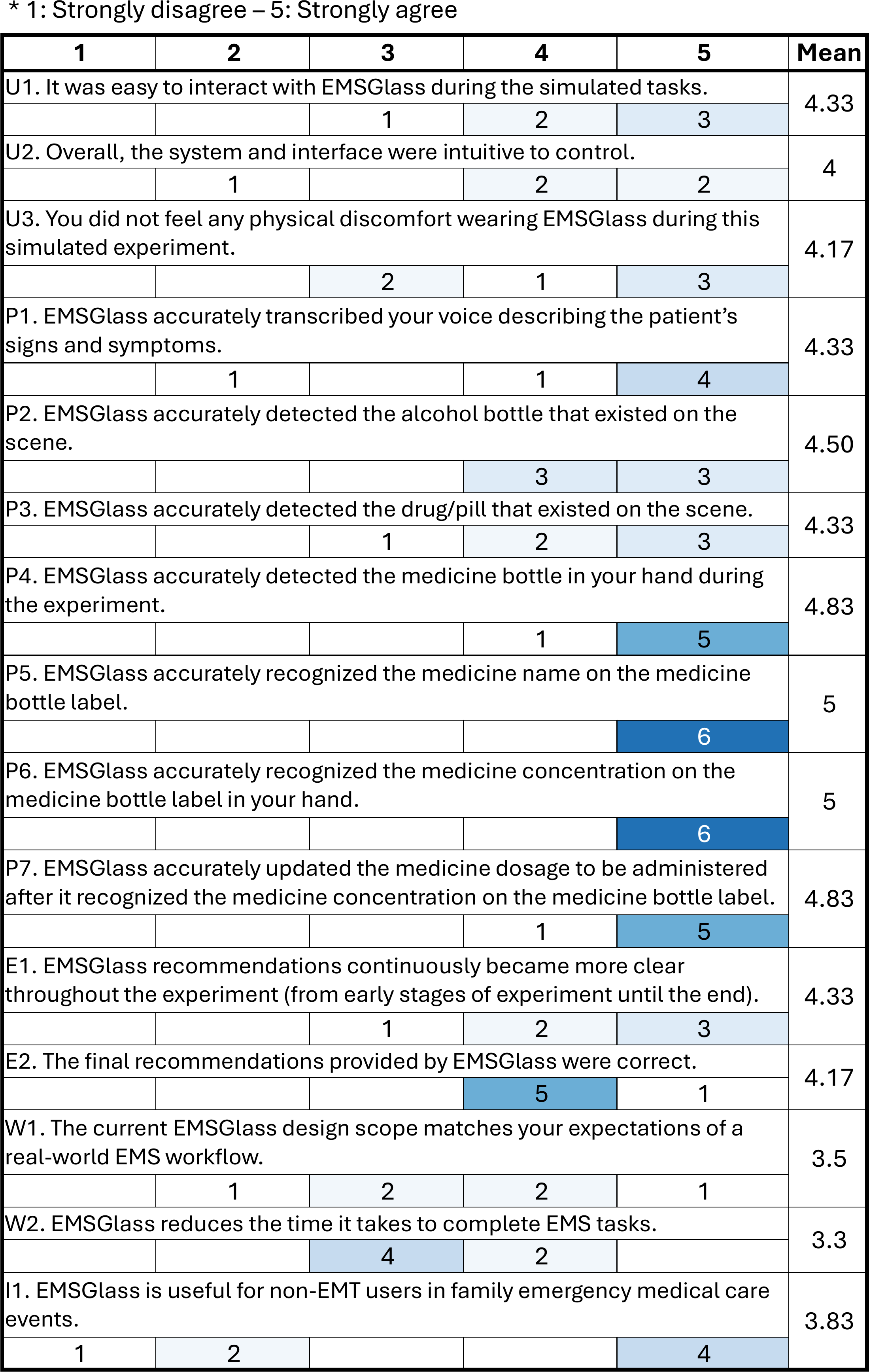}
    \caption{User study assessment scores}
     \Description{User study assessment scores}
    \label{fig:user_study_quant_scores}
\end{figure}

\section{Full evaluation of EMSNet on the 3-modal dataset}
\label{app:full_three_modal_evaluation}

As discussed in Section~\ref{sec:three_modal_backbone_evaluate}, Table~\ref{tab:multimodal_unimodal_d2} shows the full details of evaluating the EMSNet on 3-modal dataset D2 (text, vitals, image).  

\begin{table*}[]
\centering
\caption{Compare EMSGlass's multimodal models with SOTA unimodal models in tasks 1-3 on the dataset D2 (text, vitals, image).}
\label{tab:multimodal_unimodal_d2}
\scalebox{0.52}{
\begin{tabular}{llcclcccccccccccc}
\hline
 &
   &
  \multicolumn{3}{c}{Backbone(Text/Vitals/Image)} &
  \multicolumn{4}{c}{Task1-Protocol Selection (top-1/3/5)} &
  \multicolumn{4}{c}{Task2-Medicine Type Prescription (top-1/3/5)} &
  \multicolumn{4}{c}{Task3-Medicine Quant. Prescription (mse/pearsonr/spearmanr)} \\
\multirow{-2}{*}{} &
  \multirow{-2}{*}{} &
  \multicolumn{1}{l}{T} &
  \multicolumn{1}{l}{V} &
  I &
  \multicolumn{1}{c}{P} &
  \multicolumn{1}{c}{P-M} &
  \multicolumn{1}{c}{P-Q} &
  \multicolumn{1}{c}{P-M-Q} &
  \multicolumn{1}{c}{M} &
  \multicolumn{1}{c}{P-M} &
  \multicolumn{1}{c}{M-Q} &
  \multicolumn{1}{c}{P-M-Q} &
  \multicolumn{1}{c}{Q} &
  \multicolumn{1}{c}{P-Q} &
  \multicolumn{1}{c}{M-Q} &
  \multicolumn{1}{c}{P-M-Q} \\ \hline
 &
   &
  \multicolumn{1}{l}{} &
  \multicolumn{1}{l}{LSTM} &
   &
  \multicolumn{1}{c}{{\color[HTML]{9B9B9B} 0.09/0.40/0.67}} &
  \multicolumn{1}{c}{{\color[HTML]{9B9B9B} 0.09/0.24/0.59}} &
  \multicolumn{1}{c}{{\color[HTML]{9B9B9B} 0.10/0.23/0.48}} &
  \multicolumn{1}{c}{{\color[HTML]{9B9B9B} 0.09/0.24/0.51}} &
  \multicolumn{1}{c}{{\color[HTML]{9B9B9B} 0.42/0.66/0.85}} &
  \multicolumn{1}{c}{{\color[HTML]{9B9B9B} 0.39/0.60/0.86}} &
  \multicolumn{1}{c}{{\color[HTML]{9B9B9B} 0.42/0.60/0.83}} &
  \multicolumn{1}{c}{{\color[HTML]{9B9B9B} 0.39/0.51/0.82}} &
  \multicolumn{1}{c}{{\color[HTML]{9B9B9B} 1.94/0.14/0.11}} &
  \multicolumn{1}{c}{{\color[HTML]{9B9B9B} 1.96/0.17/0.12}} &
  \multicolumn{1}{c}{{\color[HTML]{9B9B9B} 1.95/0.13/0.10}} &
  \multicolumn{1}{c}{{\color[HTML]{9B9B9B} 1.97/0.17/0.10}} \\
 &
   &
  \multicolumn{1}{l}{} &
  \multicolumn{1}{l}{RNN} &
   &
  \multicolumn{1}{c}{{\color[HTML]{9B9B9B} 0.20/0.51/0.67}} &
  \multicolumn{1}{c}{{\color[HTML]{9B9B9B} 0.14/0.46/0.61}} &
  \multicolumn{1}{c}{{\color[HTML]{9B9B9B} 0.12/0.40/0.57}} &
  \multicolumn{1}{c}{{\color[HTML]{9B9B9B} 0.11/0.40/0.57}} &
  \multicolumn{1}{c}{{\color[HTML]{9B9B9B} 0.41/0.71/0.84}} &
  \multicolumn{1}{c}{{\color[HTML]{9B9B9B} 0.40/0.72/0.84}} &
  \multicolumn{1}{c}{{\color[HTML]{9B9B9B} 0.42/0.70/0.85}} &
  \multicolumn{1}{c}{{\color[HTML]{9B9B9B} 0.39/0.69/0.84}} &
  \multicolumn{1}{c}{{\color[HTML]{9B9B9B} 1.90/0.16/0.12}} &
  \multicolumn{1}{c}{{\color[HTML]{9B9B9B} 1.93/0.20/0.14}} &
  \multicolumn{1}{c}{{\color[HTML]{9B9B9B} 1.95/0.13/0.12}} &
  \multicolumn{1}{c}{{\color[HTML]{9B9B9B} 1.97/0.17/0.13}} \\
 &
   &
  \multicolumn{1}{l}{} &
  \multicolumn{1}{l}{GRU} &
   &
  \multicolumn{1}{c}{{\color[HTML]{9B9B9B} 0.48/0.68/0.74}} &
  \multicolumn{1}{c}{{\color[HTML]{9B9B9B} 0.43/0.66/0.75}} &
  \multicolumn{1}{c}{{\color[HTML]{9B9B9B} 0.43/0.66/0.75}} &
  \multicolumn{1}{c}{{\color[HTML]{9B9B9B} 0.11/0.38/0.61}} &
  \multicolumn{1}{c}{{\color[HTML]{9B9B9B} 0.43/0.92/0.94}} &
  \multicolumn{1}{c}{{\color[HTML]{9B9B9B} 0.45/0.91/0.95}} &
  \multicolumn{1}{c}{{\color[HTML]{9B9B9B} 0.45/0.92/0.95}} &
  \multicolumn{1}{c}{{\color[HTML]{9B9B9B} 0.39/0.68/0.84}} &
  \multicolumn{1}{c}{{\color[HTML]{9B9B9B} 1.93/0.03/0.00}} &
  \multicolumn{1}{c}{{\color[HTML]{9B9B9B} 2.03/0.06/0.01}} &
  \multicolumn{1}{c}{{\color[HTML]{9B9B9B} 1.95/-0.05/-0.02}} &
  \multicolumn{1}{c}{{\color[HTML]{9B9B9B} 1.98/0.16/0.12}} \\
 &
   &
  \multicolumn{1}{l}{BERTBase} &
  \multicolumn{1}{l}{} &
   &
  \multicolumn{1}{c}{{\color[HTML]{9B9B9B} 0.72/0.93/0.96}} &
  \multicolumn{1}{c}{{\color[HTML]{9B9B9B} 0.69/0.92/0.97}} &
  \multicolumn{1}{c}{{\color[HTML]{9B9B9B} 0.70/0.90/0.96}} &
  \multicolumn{1}{c}{0.71{\color[HTML]{9B9B9B} /}0.92{\color[HTML]{9B9B9B}/0.96}} &
  \multicolumn{1}{c}{{\color[HTML]{9B9B9B} 0.67/0.94/0.98}} &
  \multicolumn{1}{c}{{\color[HTML]{9B9B9B} 0.69/0.95/0.97}} &
  \multicolumn{1}{c}{{\color[HTML]{9B9B9B} 0.69/0.94/0.98}} &
  \multicolumn{1}{c}{{\color[HTML]{9B9B9B} 0.68/0.95/0.98}} &
  \multicolumn{1}{c}{{\color[HTML]{9B9B9B} 1.94/0.01/-0.02}} &
  \multicolumn{1}{c}{1.92{\color[HTML]{9B9B9B} /0.26/}0.27} &
  \multicolumn{1}{c}{{\color[HTML]{9B9B9B} 1.95/0.24/0.22}} &
  \multicolumn{1}{c}{{\color[HTML]{9B9B9B} 1.93/0.25/0.25}} \\
 &
   &
  \multicolumn{1}{l}{TinyBERT} &
  \multicolumn{1}{l}{} &
   &
  \multicolumn{1}{c}{{\color[HTML]{9B9B9B} 0.72/0.92/0.96}} &
  \multicolumn{1}{c}{{\color[HTML]{9B9B9B} 0.69/0.91/0.96}} &
  \multicolumn{1}{c}{{\color[HTML]{9B9B9B} 0.69/0.91/0.96}} &
  \multicolumn{1}{c}{{\color[HTML]{9B9B9B} 0.69/0.89/0.96}} &
  \multicolumn{1}{c}{{\color[HTML]{9B9B9B} 0.67/0.96/0.98}} &
  \multicolumn{1}{c}{{\color[HTML]{9B9B9B} 0.66/0.95/0.98}} &
  \multicolumn{1}{c}{{\color[HTML]{9B9B9B} 0.65/0.96/0.98}} &
  \multicolumn{1}{c}{{\color[HTML]{9B9B9B} 0.66/0.95/0.98}} &
  \multicolumn{1}{c}{{\color[HTML]{9B9B9B} 1.96/0.13/0.17}} &
  \multicolumn{1}{c}{{\color[HTML]{9B9B9B} 1.96/0.21/0.23}} &
  \multicolumn{1}{c}{{\color[HTML]{9B9B9B} 1.95/0.17/0.21}} &
  \multicolumn{1}{c}{{\color[HTML]{9B9B9B} 1.91/0.24/0.27}} \\
 &
   &
  \multicolumn{1}{l}{MobileBERT} &
  \multicolumn{1}{l}{} &
   &
  \multicolumn{1}{c}{{\color[HTML]{9B9B9B} 0.73/0.93/0.97}} &
  \multicolumn{1}{c}{{\color[HTML]{9B9B9B} 0.73/0.91/0.96}} &
  \multicolumn{1}{c}{{\color[HTML]{9B9B9B} 0.66/0.86/0.93}} &
  \multicolumn{1}{c}{{\color[HTML]{9B9B9B} 0.66/0.90/0.95}} &
  \multicolumn{1}{c}{{\color[HTML]{9B9B9B} 0.67/0.95/0.98}} &
  \multicolumn{1}{c}{{\color[HTML]{9B9B9B} 0.69/0.95/0.97}} &
  \multicolumn{1}{c}{{\color[HTML]{9B9B9B} 0.67/0.96/0.98}} &
  \multicolumn{1}{c}{{\color[HTML]{9B9B9B} 0.68/0.95/0.98}} &
  \multicolumn{1}{c}{{\color[HTML]{9B9B9B} 1.91/0.17/0.22}} &
  \multicolumn{1}{c}{{\color[HTML]{9B9B9B} 1.99/0.16/0.22}} &
  \multicolumn{1}{c}{1.89{\color[HTML]{9B9B9B} /0.20/0.24}} &
  \multicolumn{1}{c}{{\color[HTML]{9B9B9B} 1.94/0.21/0.25}} \\
\multirow{-7}{*}{\rotatebox{90}{Unimodal (SOTA)}} &
  \multirow{-7}{*}{\rotatebox{90}{No FT}} &
  \multicolumn{1}{l}{} &
  \multicolumn{1}{l}{} &
  FC &
  {\color[HTML]{9B9B9B} 0.00/0.02/0.02} &
  {\color[HTML]{9B9B9B} 0.00/0.01/0.01} &
  {\color[HTML]{9B9B9B} 0.00/0.01/0.01} &
  {\color[HTML]{9B9B9B} 0.00/0.01/0.01} &
  {\color[HTML]{9B9B9B} 0.00/0.03/0.05} &
  {\color[HTML]{9B9B9B} 0.01/0.41/0.48} &
  {\color[HTML]{9B9B9B} 0.00/0.03/0.05} &
  {\color[HTML]{9B9B9B} 0.01/0.41/0.48} &
  {\color[HTML]{9B9B9B} 2.26/0.10/0.12} &
  {\color[HTML]{9B9B9B} 2.10/0.16/0.16} &
  {\color[HTML]{9B9B9B} 1.97/-0.11/-0.12} &
  {\color[HTML]{9B9B9B} 2.08/0.16/0.16} \\ \hline
 &
   &
  \multicolumn{1}{l}{} &
  \multicolumn{1}{l}{LSTM} &
   &
  \multicolumn{1}{c}{{\color[HTML]{9B9B9B} 0.71/0.92/0.97}} &
  \multicolumn{1}{c}{{\color[HTML]{9B9B9B} 0.66/0.90/0.95}} &
  \multicolumn{1}{c}{{\color[HTML]{9B9B9B} 0.10/0.27/0.52}} &
  \multicolumn{1}{c}{{\color[HTML]{9B9B9B} 0.70/0.91/0.96}} &
  \multicolumn{1}{c}{{\color[HTML]{9B9B9B} 0.71/0.96/0.99}} &
  \multicolumn{1}{c}{{\color[HTML]{9B9B9B} 0.70/0.94/0.98}} &
  \multicolumn{1}{c}{0.73{\color[HTML]{9B9B9B} /0.96/0.97}} &
  \multicolumn{1}{c}{{\color[HTML]{9B9B9B} 0.68/0.94/0.97}} &
  \multicolumn{1}{c}{{\color[HTML]{9B9B9B} 1.94/0.13/0.10}} &
  \multicolumn{1}{c}{{\color[HTML]{9B9B9B} 1.97/0.17/0.12}} &
  \multicolumn{1}{c}{{\color[HTML]{9B9B9B} 1.91/}0.26{\color[HTML]{9B9B9B}/}0.25} &
  \multicolumn{1}{c}{{\color[HTML]{9B9B9B} 1.98/0.27/0.25}} \\
 &
   &
  \multicolumn{1}{l}{} &
  \multicolumn{1}{l}{RNN} &
   &
  \multicolumn{1}{c}{{\color[HTML]{9B9B9B} 0.72/0.94/}0.98} &
  \multicolumn{1}{c}{{\color[HTML]{9B9B9B} 0.68/0.91/0.96}} &
  \multicolumn{1}{c}{{\color[HTML]{9B9B9B} 0.70/0.90/}0.97} &
  \multicolumn{1}{c}{{\color[HTML]{9B9B9B} 0.70/0.91/}0.97} &
  \multicolumn{1}{c}{{\color[HTML]{9B9B9B} 0.68/0.95/0.98}} &
  \multicolumn{1}{c}{{\color[HTML]{9B9B9B} 0.69/0.95/0.98}} &
  \multicolumn{1}{c}{{\color[HTML]{9B9B9B} 0.43/0.68/0.84}} &
  \multicolumn{1}{c}{0.70{\color[HTML]{9B9B9B} /0.95/0.98}} &
  \multicolumn{1}{c}{{\color[HTML]{9B9B9B} 1.90/0.16/0.12}} &
  \multicolumn{1}{c}{{\color[HTML]{9B9B9B} 2.00/0.24/0.24}} &
  \multicolumn{1}{c}{{\color[HTML]{9B9B9B} 1.96/0.11/0.10}} &
  \multicolumn{1}{c}{1.88{\color[HTML]{9B9B9B} /}0.27{\color[HTML]{9B9B9B}/}0.28} \\
 &
   &
  \multicolumn{1}{l}{\multirow{-3}{*}{BERTBase}} &
  \multicolumn{1}{l}{GRU} &
   &
  \multicolumn{1}{c}{{\color[HTML]{9B9B9B} 0.72/0.92/0.97}} &
  \multicolumn{1}{c}{{\color[HTML]{9B9B9B} 0.72/0.92/0.97}} &
  \multicolumn{1}{c}{{\color[HTML]{9B9B9B} 0.70/}0.92{\color[HTML]{9B9B9B}/0.96}} &
  \multicolumn{1}{c}{{\color[HTML]{9B9B9B} 0.66/0.89/0.94}} &
  \multicolumn{1}{c}{{\color[HTML]{9B9B9B} 0.66/0.95/0.98}} &
  \multicolumn{1}{c}{{\color[HTML]{9B9B9B} 0.70/0.95/0.98}} &
  \multicolumn{1}{c}{{\color[HTML]{9B9B9B} 0.69/0.95/0.99}} &
  \multicolumn{1}{c}{{\color[HTML]{9B9B9B} 0.68/0.93/0.98}} &
  \multicolumn{1}{c}{{\color[HTML]{9B9B9B} 1.95/0.13/0.13}} &
  \multicolumn{1}{c}{{\color[HTML]{9B9B9B} 1.95/}0.26{\color[HTML]{9B9B9B}/0.24}} &
  \multicolumn{1}{c}{{\color[HTML]{9B9B9B} 1.93/0.22/0.22}} &
  \multicolumn{1}{c}{{\color[HTML]{9B9B9B} 1.94/0.24/0.24}} \\
 &
   &
  \multicolumn{1}{l}{} &
  \multicolumn{1}{l}{LSTM} &
   &
  \multicolumn{1}{c}{0.74{\color[HTML]{9B9B9B} /0.93/0.96}} &
  \multicolumn{1}{c}{{\color[HTML]{9B9B9B} 0.69/0.91/0.95}} &
  \multicolumn{1}{c}{{\color[HTML]{9B9B9B} 0.66/0.90/0.95}} &
  \multicolumn{1}{c}{{\color[HTML]{9B9B9B} 0.67/0.91/0.95}} &
  \multicolumn{1}{c}{{\color[HTML]{9B9B9B} 0.72/0.95/0.98}} &
  \multicolumn{1}{c}{{\color[HTML]{9B9B9B} 0.66/0.95/0.97}} &
  \multicolumn{1}{c}{{\color[HTML]{9B9B9B} 0.69/0.96/0.97}} &
  \multicolumn{1}{c}{{\color[HTML]{9B9B9B} 0.67/0.95/0.98}} &
  \multicolumn{1}{c}{{\color[HTML]{9B9B9B} 1.94/0.20/0.22}} &
  \multicolumn{1}{c}{{\color[HTML]{9B9B9B} 1.94/0.23/0.23}} &
  \multicolumn{1}{c}{{\color[HTML]{9B9B9B} 1.95/0.21/0.23}} &
  \multicolumn{1}{c}{{\color[HTML]{9B9B9B} 1.91/0.25/0.26}} \\
 &
   &
  \multicolumn{1}{l}{} &
  \multicolumn{1}{l}{RNN} &
   &
  \multicolumn{1}{c}{{\color[HTML]{9B9B9B} 0.71/0.93/0.95}} &
  \multicolumn{1}{c}{{\color[HTML]{9B9B9B} 0.67/0.91/0.95}} &
  \multicolumn{1}{c}{{\color[HTML]{9B9B9B} 0.69/0.91/0.96}} &
  \multicolumn{1}{c}{{\color[HTML]{9B9B9B} 0.68/0.91/0.95}} &
  \multicolumn{1}{c}{{\color[HTML]{9B9B9B} 0.69/0.95/0.99}} &
  \multicolumn{1}{c}{{\color[HTML]{9B9B9B} 0.68/}0.96{\color[HTML]{9B9B9B}/0.98}} &
  \multicolumn{1}{c}{{\color[HTML]{9B9B9B} 0.67/0.96/0.98}} &
  \multicolumn{1}{c}{{\color[HTML]{9B9B9B} 0.67/0.95/0.98}} &
  \multicolumn{1}{c}{{\color[HTML]{9B9B9B} 1.91/0.20/0.23}} &
  \multicolumn{1}{c}{{\color[HTML]{9B9B9B} 1.95/0.24/0.25}} &
  \multicolumn{1}{c}{{\color[HTML]{9B9B9B} 1.92/0.22/0.23}} &
  \multicolumn{1}{c}{{\color[HTML]{9B9B9B} 1.95/0.23/0.24}} \\
 &
   &
  \multicolumn{1}{l}{\multirow{-3}{*}{TinyBERT}} &
  \multicolumn{1}{l}{GRU} &
   &
  \multicolumn{1}{c}{{\color[HTML]{9B9B9B} 0.72/0.92/0.96}} &
  \multicolumn{1}{c}{{\color[HTML]{9B9B9B} 0.68/0.91/0.95}} &
  \multicolumn{1}{c}{{\color[HTML]{9B9B9B} 0.69/0.90/0.96}} &
  \multicolumn{1}{c}{{\color[HTML]{9B9B9B} 0.68/0.91/0.95}} &
  \multicolumn{1}{c}{{\color[HTML]{9B9B9B} 0.69/0.95/0.98}} &
  \multicolumn{1}{c}{{\color[HTML]{9B9B9B} 0.68/0.95/0.98}} &
  \multicolumn{1}{c}{{\color[HTML]{9B9B9B} 0.70/0.95/}0.99} &
  \multicolumn{1}{c}{{\color[HTML]{9B9B9B} 0.68/0.94/0.98}} &
  \multicolumn{1}{c}{{\color[HTML]{9B9B9B} 1.97/0.18/0.21}} &
  \multicolumn{1}{c}{{\color[HTML]{9B9B9B} 1.97/0.22/0.24}} &
  \multicolumn{1}{c}{{\color[HTML]{9B9B9B} 1.98/0.18/0.20}} &
  \multicolumn{1}{c}{{\color[HTML]{9B9B9B} 1.97/0.22/0.22}} \\
 &
   &
  \multicolumn{1}{l}{} &
  \multicolumn{1}{l}{LSTM} &
   &
  \multicolumn{1}{c}{{\color[HTML]{9B9B9B} 0.72/0.93/0.96}} &
  \multicolumn{1}{c}{{\color[HTML]{9B9B9B} 0.71/0.91/0.95}} &
  \multicolumn{1}{c}{{\color[HTML]{9B9B9B} 0.66/0.88/0.95}} &
  \multicolumn{1}{c}{{\color[HTML]{9B9B9B} 0.68/0.89/0.95}} &
  \multicolumn{1}{c}{{\color[HTML]{9B9B9B} 0.68/0.95/0.98}} &
  \multicolumn{1}{c}{{\color[HTML]{9B9B9B} 0.69/0.95/0.98}} &
  \multicolumn{1}{c}{{\color[HTML]{9B9B9B} 0.69/0.95/0.98}} &
  \multicolumn{1}{c}{{\color[HTML]{9B9B9B} 0.67/0.95/0.98}} &
  \multicolumn{1}{c}{{\color[HTML]{9B9B9B} 1.93/0.19/0.20}} &
  \multicolumn{1}{c}{{\color[HTML]{9B9B9B} 1.95/0.21/0.23}} &
  \multicolumn{1}{c}{{\color[HTML]{9B9B9B} 1.92/0.21/0.24}} &
  \multicolumn{1}{c}{{\color[HTML]{9B9B9B} 1.94/0.22/0.24}} \\
 &
   &
  \multicolumn{1}{l}{} &
  \multicolumn{1}{l}{RNN} &
   &
  \multicolumn{1}{c}{{\color[HTML]{9B9B9B} 0.72/}0.94{\color[HTML]{9B9B9B}/0.97}} &
  \multicolumn{1}{c}{0.73{\color[HTML]{9B9B9B} /0.91/}0.97} &
  \multicolumn{1}{c}{{\color[HTML]{9B9B9B} 0.67/0.89/0.96}} &
  \multicolumn{1}{c}{{\color[HTML]{9B9B9B} 0.66/0.88/0.95}} &
  \multicolumn{1}{c}{{\color[HTML]{9B9B9B} 0.71/0.94/0.98}} &
  \multicolumn{1}{c}{{\color[HTML]{9B9B9B} 0.70/0.95/0.97}} &
  \multicolumn{1}{c}{{\color[HTML]{9B9B9B} 0.67/0.95/0.98}} &
  \multicolumn{1}{c}{{\color[HTML]{9B9B9B} 0.66/0.95/0.98}} &
  \multicolumn{1}{c}{{\color[HTML]{9B9B9B} 1.92/}0.20{\color[HTML]{9B9B9B}/}0.23} &
  \multicolumn{1}{c}{{\color[HTML]{9B9B9B} 1.95/0.22/0.25}} &
  \multicolumn{1}{c}{{\color[HTML]{9B9B9B} 1.93/0.20/0.22}} &
  \multicolumn{1}{c}{{\color[HTML]{9B9B9B} 2.00/0.17/0.21}} \\
 &
  \multirow{-9}{*}{\rotatebox{90}{No FT}} &
  \multicolumn{1}{l}{\multirow{-3}{*}{MobileBERT}} &
  \multicolumn{1}{l}{GRU} &
   &
  \multicolumn{1}{c}{{\color[HTML]{9B9B9B} 0.72/0.92/0.97}} &
  \multicolumn{1}{c}{{\color[HTML]{9B9B9B} 0.70/}0.92{\color[HTML]{9B9B9B}/0.96}} &
  \multicolumn{1}{c}{{\color[HTML]{9B9B9B} 0.65/0.89/0.95}} &
  \multicolumn{1}{c}{{\color[HTML]{9B9B9B} 0.69/0.90/0.95}} &
  \multicolumn{1}{c}{0.72{\color[HTML]{9B9B9B} /0.95/}0.99} &
  \multicolumn{1}{c}{{\color[HTML]{9B9B9B} 0.69/0.94/}0.99} &
  \multicolumn{1}{c}{{\color[HTML]{9B9B9B} 0.68/0.96/0.98}} &
  \multicolumn{1}{c}{{\color[HTML]{9B9B9B} 0.68/0.95/0.98}} &
  \multicolumn{1}{c}{{\color[HTML]{9B9B9B} 1.93/0.19/0.22}} &
  \multicolumn{1}{c}{{\color[HTML]{9B9B9B} 1.94/0.22/0.23}} &
  \multicolumn{1}{c}{{\color[HTML]{9B9B9B} 1.91/0.21/0.22}} &
  \multicolumn{1}{c}{{\color[HTML]{9B9B9B} 1.96/0.22/0.23}} \\ \cline{2-17} 
 &
   &
  \multicolumn{1}{l}{} &
  LSTM &
  FC &
  {\color[HTML]{9B9B9B} 0.48/0.68/0.77} &
  {\color[HTML]{9B9B9B} 0.43/0.66/0.75} &
  {\color[HTML]{9B9B9B} 0.43/0.66/0.75} &
  {\color[HTML]{9B9B9B} 0.43/0.66/0.75} &
  {\color[HTML]{9B9B9B} 0.43/0.92/0.93} &
  {\color[HTML]{9B9B9B} 0.45/0.91/0.95} &
  {\color[HTML]{9B9B9B} 0.43/0.92/0.93} &
  {\color[HTML]{9B9B9B} 0.45/0.91/0.95} &
  {\color[HTML]{9B9B9B} 1.93/0.09/0.03} &
  {\color[HTML]{9B9B9B} 2.00/0.02/0.04} &
  {\color[HTML]{9B9B9B} 2.07/0.03/0.01} &
  {\color[HTML]{9B9B9B} 2.00/0.09/0.07} \\
 &
   &
  \multicolumn{1}{l}{} &
  RNN &
  FC &
  {\color[HTML]{9B9B9B} 0.48/0.68/0.77} &
  {\color[HTML]{9B9B9B} 0.43/0.66/0.74} &
  {\color[HTML]{9B9B9B} 0.43/0.66/0.72} &
  {\color[HTML]{9B9B9B} 0.43/0.66/0.72} &
  {\color[HTML]{9B9B9B} 0.43/0.92/0.95} &
  {\color[HTML]{9B9B9B} 0.45/0.91/0.93} &
  {\color[HTML]{9B9B9B} 0.43/0.92/0.95} &
  {\color[HTML]{9B9B9B} 0.45/0.91/0.93} &
  {\color[HTML]{9B9B9B} 1.93/0.06/0.06} &
  {\color[HTML]{9B9B9B} 1.99/0.13/0.06} &
  {\color[HTML]{9B9B9B} 1.94/0.07/0.06} &
  {\color[HTML]{9B9B9B} 2.00/0.06/0.01} \\
 &
   &
  \multicolumn{1}{l}{} &
  GRU &
  FC &
  {\color[HTML]{9B9B9B} 0.48/0.68/0.77} &
  {\color[HTML]{9B9B9B} 0.43/0.66/0.75} &
  {\color[HTML]{9B9B9B} 0.43/0.66/0.75} &
  {\color[HTML]{9B9B9B} 0.43/0.66/0.75} &
  {\color[HTML]{9B9B9B} 0.43/0.92/0.93} &
  {\color[HTML]{9B9B9B} 0.45/0.91/0.93} &
  {\color[HTML]{9B9B9B} 0.43/0.92/0.93} &
  {\color[HTML]{9B9B9B} 0.45/0.91/0.94} &
  {\color[HTML]{9B9B9B} 1.94/-0.02/-0.01} &
  {\color[HTML]{9B9B9B} 2.00/0.11/0.07} &
  {\color[HTML]{9B9B9B} 1.95/-0.02/-0.03} &
  {\color[HTML]{9B9B9B} 2.00/0.01/0.01} \\
 &
   &
  BERTBase &
  \multicolumn{1}{l}{} &
  FC &
  {\color[HTML]{9B9B9B} 0.68/0.88/0.93} &
  {\color[HTML]{9B9B9B} 0.67/0.87/0.92} &
  {\color[HTML]{9B9B9B} 0.52/0.75/0.82} &
  {\color[HTML]{9B9B9B} 0.65/0.83/0.89} &
  {\color[HTML]{9B9B9B} 0.68/0.94/0.97} &
  {\color[HTML]{9B9B9B} 0.66/0.93/0.97} &
  {\color[HTML]{9B9B9B} 0.65/0.95/0.97} &
  {\color[HTML]{9B9B9B} 0.65/0.92/0.96} &
  {\color[HTML]{9B9B9B} 2.01/0.09/0.14} &
  {\color[HTML]{9B9B9B} 1.97/0.14/0.17} &
  {\color[HTML]{9B9B9B} 2.10/0.13/0.15} &
  {\color[HTML]{9B9B9B} 2.55/0.12/0.14} \\
 &
   &
  TinyBERT &
  \multicolumn{1}{l}{} &
  FC &
  {\color[HTML]{9B9B9B} 0.63/0.80/0.88} &
  {\color[HTML]{9B9B9B} 0.58/0.70/0.79} &
  {\color[HTML]{9B9B9B} 0.43/0.66/0.73} &
  {\color[HTML]{9B9B9B} 0.51/0.69/0.79} &
  {\color[HTML]{9B9B9B} 0.66/0.94/0.97} &
  {\color[HTML]{9B9B9B} 0.67/0.92/0.95} &
  {\color[HTML]{9B9B9B} 0.67/0.93/0.96} &
  {\color[HTML]{9B9B9B} 0.66/0.90/0.95} &
  {\color[HTML]{9B9B9B} 1.87/0.19/0.20} &
  {\color[HTML]{9B9B9B} 1.96/0.15/0.18} &
  {\color[HTML]{9B9B9B} 1.91/0.17/0.17} &
  {\color[HTML]{9B9B9B} 1.98/0.16/0.18} \\
 &
  \multirow{-6}{*}{\rotatebox{90}{FT w/o PMI}} &
  MobileBERT &
  \multicolumn{1}{l}{} &
  FC &
  {\color[HTML]{9B9B9B} 0.49/0.67/0.72} &
  {\color[HTML]{9B9B9B} 0.42/0.60/0.65} &
  {\color[HTML]{9B9B9B} 0.16/0.59/0.64} &
  {\color[HTML]{9B9B9B} 0.43/0.47/0.63} &
  {\color[HTML]{9B9B9B} 0.51/0.89/0.93} &
  {\color[HTML]{9B9B9B} 0.44/0.88/0.91} &
  {\color[HTML]{9B9B9B} 0.41/0.46/0.48} &
  {\color[HTML]{9B9B9B} 0.45/0.48/0.54} &
  {\color[HTML]{9B9B9B} 6e2/-0.04/0.01} &
  {\color[HTML]{9B9B9B} 1e6/0.06/0.06} &
  {\color[HTML]{9B9B9B} 2e5/0.03/0.04} &
  {\color[HTML]{9B9B9B} 10e6/0.05/0.07} \\ \cline{2-17}
 &
   &
  \multicolumn{1}{l}{} &
  LSTM &
  FC &
  {\color[HTML]{9B9B9B} 0.48/0.68/0.76} &
  {\color[HTML]{9B9B9B} 0.43/0.65/0.75} &
  {\color[HTML]{9B9B9B} 0.43/0.65/0.74} &
  {\color[HTML]{9B9B9B} 0.43/0.66/0.76} &
  {\color[HTML]{9B9B9B} 0.50/0.90/0.94} &
  {\color[HTML]{9B9B9B} 0.40/0.91/0.93} &
  {\color[HTML]{9B9B9B} 0.43/0.90/0.94} &
  {\color[HTML]{9B9B9B} 0.49/0.88/0.95} &
  {\color[HTML]{9B9B9B} 1.93/0.09/0.06} &
  {\color[HTML]{9B9B9B} 2.00/0.02/-0.02} &
  {\color[HTML]{9B9B9B} 1.93/0.06/-0.02} &
  {\color[HTML]{9B9B9B} 1.99/0.11/0.08} \\
 &
   &
  \multicolumn{1}{l}{} &
  RNN &
  FC &
  {\color[HTML]{9B9B9B} 0.48/0.69/0.77} &
  {\color[HTML]{9B9B9B} 0.43/0.66/0.74} &
  {\color[HTML]{9B9B9B} 0.43/0.66/0.75} &
  {\color[HTML]{9B9B9B} 0.43/0.66/0.75} &
  {\color[HTML]{9B9B9B} 0.45/0.91/0.95} &
  {\color[HTML]{9B9B9B} 0.54/0.91/0.95} &
  {\color[HTML]{9B9B9B} 0.48/0.91/0.95} &
  {\color[HTML]{9B9B9B} 0.45/0.89/0.94} &
  {\color[HTML]{9B9B9B} 1.92/0.08/0.07} &
  {\color[HTML]{9B9B9B} 2.00/0.06/0.06} &
  {\color[HTML]{9B9B9B} 1.96/0.03/0.04} &
  {\color[HTML]{9B9B9B} 1.99/0.07/0.09} \\
 &
   &
  \multicolumn{1}{l}{} &
  GRU &
  FC &
  {\color[HTML]{9B9B9B} 0.48/0.68/0.74} &
  {\color[HTML]{9B9B9B} 0.43/0.66/0.75} &
  {\color[HTML]{9B9B9B} 0.43/0.65/0.75} &
  {\color[HTML]{9B9B9B} 0.43/0.66/0.73} &
  {\color[HTML]{9B9B9B} 0.50/0.92/0.95} &
  {\color[HTML]{9B9B9B} 0.48/0.91/0.95} &
  {\color[HTML]{9B9B9B} 0.50/0.92/0.95} &
  {\color[HTML]{9B9B9B} 0.40/0.88/0.94} &
  {\color[HTML]{9B9B9B} 1.91/0.12/0.12} &
  {\color[HTML]{9B9B9B} 1.97/0.11/0.08} &
  {\color[HTML]{9B9B9B} 1.92/0.09/0.03} &
  {\color[HTML]{9B9B9B} 2.00/0.05/0.05} \\
 &
   &
  BERTBase &
  \multicolumn{1}{l}{} &
  FC &
  {\color[HTML]{9B9B9B} 0.69/0.90/0.94} &
  {\color[HTML]{9B9B9B} 0.69/0.88/0.94} &
  {\color[HTML]{9B9B9B} 0.66/0.89/0.93} &
  {\color[HTML]{9B9B9B} 0.70/0.89/0.93} &
  {\color[HTML]{9B9B9B} 0.65/0.95/0.98} &
  {\color[HTML]{9B9B9B} 0.68/0.94/0.98} &
  {\color[HTML]{9B9B9B} 0.64/0.96/0.98} &
  {\color[HTML]{9B9B9B} 0.67/0.94/0.97} &
  {\color[HTML]{9B9B9B} 1.94/0.10/0.10} &
  {\color[HTML]{9B9B9B} 2.31/0.20/0.22} &
  {\color[HTML]{9B9B9B} 1.95/0.21/0.22} &
  {\color[HTML]{9B9B9B} 2.53/0.17/0.16} \\
 &
   &
  TinyBERT &
  \multicolumn{1}{l}{} &
  FC &
  {\color[HTML]{9B9B9B} 0.67/0.86/0.91} &
  {\color[HTML]{9B9B9B} 0.64/0.81/0.87} &
  {\color[HTML]{9B9B9B} 0.56/0.81/0.86} &
  {\color[HTML]{9B9B9B} 0.58/0.78/0.85} &
  {\color[HTML]{9B9B9B} 0.66/0.95/0.97} &
  {\color[HTML]{9B9B9B} 0.68/0.94/0.97} &
  {\color[HTML]{9B9B9B} 0.68/0.95/0.97} &
  {\color[HTML]{9B9B9B} 0.68/0.94/0.98} &
  {\color[HTML]{9B9B9B} 1.91/0.17/0.20} &
  {\color[HTML]{9B9B9B} 2.02/0.17/0.20} &
  {\color[HTML]{9B9B9B} 1.97/0.16/0.18} &
  {\color[HTML]{9B9B9B} 2.01/0.16/0.18} \\
\multirow{-21}{*}{\rotatebox{90}{\textbf{Multimodal: 2-modal (our)}}} &
  \multirow{-6}{*}{\rotatebox{90}{\textbf{FT w/ PMI}}} &
  MobileBERT &
  \multicolumn{1}{l}{} &
  FC &
  {\color[HTML]{9B9B9B} 0.71/0.92/0.95} &
  {\color[HTML]{9B9B9B} 0.69/0.90/0.93} &
  {\color[HTML]{9B9B9B} 0.63/0.83/0.88} &
  {\color[HTML]{9B9B9B} 0.69/0.88/0.94} &
  {\color[HTML]{9B9B9B} 0.69/0.94/0.97} &
  {\color[HTML]{9B9B9B} 0.68/0.95/0.98} &
  {\color[HTML]{9B9B9B} 0.68/0.95/0.98} &
  {\color[HTML]{9B9B9B} 0.68/0.94/0.97} &
  {\color[HTML]{9B9B9B} 2.05/0.13/0.16} &
  {\color[HTML]{9B9B9B} 2.18/0.08/0.11} &
  {\color[HTML]{9B9B9B} 1.97/0.12/0.15} &
  {\color[HTML]{9B9B9B} 2.20/0.14/0.18} \\ \cline{2-17} 
 &
   &
   &
  LSTM &
  FC &
  {\color[HTML]{9B9B9B} 0.69/0.89/0.93} &
  {\color[HTML]{9B9B9B} 0.68/0.88/0.92} &
  {\color[HTML]{9B9B9B} 0.64/0.83/0.86} &
  {\color[HTML]{9B9B9B} 0.67/0.86/0.89} &
  {\color[HTML]{9B9B9B} 0.66/0.93/0.98} &
  {\color[HTML]{9B9B9B} 0.68/0.93/0.97} &
  {\color[HTML]{9B9B9B} 0.6//0.94/0.96} &
  {\color[HTML]{9B9B9B} 0.69/0.95/0.97} &
  {\color[HTML]{9B9B9B} 1.94/0.04/-0.08} &
  {\color[HTML]{9B9B9B} 2.35/0.10/0.17} &
  {\color[HTML]{9B9B9B} 2.05/0.15/0.16} &
  {\color[HTML]{9B9B9B} 2.42/0.11/0.18} \\
 &
   &
   &
  RNN &
  FC &
  {\color[HTML]{9B9B9B} 0.67/0.86/0.91} &
  {\color[HTML]{9B9B9B} 0.67/0.87/0.91} &
  {\color[HTML]{9B9B9B} 0.67/0.86/0.91} &
  {\color[HTML]{9B9B9B} 0.67/0.87/0.90} &
  {\color[HTML]{9B9B9B} 0.67/0.94/0.98} &
  {\color[HTML]{9B9B9B} 0.67/0.94/0.98} &
  {\color[HTML]{9B9B9B} 0.64/0.92/0.96} &
  {\color[HTML]{9B9B9B} 0.67/0.93/0.98} &
  {\color[HTML]{9B9B9B} 1.90/0.15/0.21} &
  {\color[HTML]{9B9B9B} 2.05/0.20/0.23} &
  {\color[HTML]{9B9B9B} 1.93/0.13/0.16} &
  {\color[HTML]{9B9B9B} 2.50/0.13/0.16} \\
 &
   &
  \multirow{-3}{*}{BERTBase} &
  GRU &
  FC &
  {\color[HTML]{9B9B9B} 0.69/0.89/0.93} &
  {\color[HTML]{9B9B9B} 0.67/0.89/0.93} &
  {\color[HTML]{9B9B9B} 0.66/0.84/0.89} &
  {\color[HTML]{9B9B9B} 0.65/0.86/0.89} &
  {\color[HTML]{9B9B9B} 0.62/0.93/0.96} &
  {\color[HTML]{9B9B9B} 0.68/0.94/0.98} &
  {\color[HTML]{9B9B9B} 0.67/0.93/0.97} &
  {\color[HTML]{9B9B9B} 0.67/0.94/0.97} &
  {\color[HTML]{9B9B9B} 1.92/0.09/0.09} &
  {\color[HTML]{9B9B9B} 2.01/0.19/0.25} &
  {\color[HTML]{9B9B9B} 2.18/0.18/0.20} &
  {\color[HTML]{9B9B9B} 2.15/0.16/0.24} \\
 &
   &
   &
  LSTM &
  FC &
  {\color[HTML]{9B9B9B} 0.65/0.82/0.89} &
  {\color[HTML]{9B9B9B} 0.59/0.78/0.84} &
  {\color[HTML]{9B9B9B} 0.55/0.77/0.82} &
  {\color[HTML]{9B9B9B} 0.54/0.76/0.84} &
  {\color[HTML]{9B9B9B} 0.66/0.93/0.97} &
  {\color[HTML]{9B9B9B} 0.66/0.93/0.96} &
  {\color[HTML]{9B9B9B} 0.66/0.93/0.97} &
  {\color[HTML]{9B9B9B} 0.67/0.94/0.96} &
  {\color[HTML]{9B9B9B} 1.90/0.16/0.17} &
  {\color[HTML]{9B9B9B} 1.94/0.20/0.22} &
  {\color[HTML]{9B9B9B} 1.97/0.14/0.16} &
  {\color[HTML]{9B9B9B} 1.95/0.19/0.21} \\
 &
   &
   &
  RNN &
  FC &
  {\color[HTML]{9B9B9B} 0.63/0.83/0.89} &
  {\color[HTML]{9B9B9B} 0.61/0.80/0.84} &
  {\color[HTML]{9B9B9B} 0.55/0.74/0.82} &
  {\color[HTML]{9B9B9B} 0.53/0.76/0.82} &
  {\color[HTML]{9B9B9B} 0.65/0.95/0.97} &
  {\color[HTML]{9B9B9B} 0.65/0.92/0.97} &
  {\color[HTML]{9B9B9B} 0.66/0.92/0.95} &
  {\color[HTML]{9B9B9B} 0.65/0.92/0.96} &
  1.87{\color[HTML]{9B9B9B} /0.18/0.18} &
  {\color[HTML]{9B9B9B} 1.98/0.17/0.19} &
  {\color[HTML]{9B9B9B} 1.92/0.13/0.17} &
  {\color[HTML]{9B9B9B} 1.93/0.19/0.20} \\
 &
   &
  \multirow{-3}{*}{TinyBERT} &
  GRU &
  FC &
  {\color[HTML]{9B9B9B} 0.63/0.82/0.88} &
  {\color[HTML]{9B9B9B} 0.61/0.81/0.85} &
  {\color[HTML]{9B9B9B} 0.52/0.74/0.83} &
  {\color[HTML]{9B9B9B} 0.54/0.74/0.81} &
  {\color[HTML]{9B9B9B} 0.64/0.94/0.97} &
  {\color[HTML]{9B9B9B} 0.68/0.93/0.97} &
  {\color[HTML]{9B9B9B} 0.66/0.93/0.97} &
  {\color[HTML]{9B9B9B} 0.67/0.93/0.97} &
  {\color[HTML]{9B9B9B} 1.94/0.14/0.16} &
  {\color[HTML]{9B9B9B} 1.97/0.18/0.21} &
  {\color[HTML]{9B9B9B} 1.92/0.17/0.19} &
  {\color[HTML]{9B9B9B} 1.93/0.20/0.21} \\
 &
   &
   &
  LSTM &
  FC &
  {\color[HTML]{9B9B9B} 0.49/0.72/0.84} &
  {\color[HTML]{9B9B9B} 0.44/0.68/0.77} &
  {\color[HTML]{9B9B9B} 0.02/0.09/0.11} &
  {\color[HTML]{9B9B9B} 0.02/0.05/0.09} &
  {\color[HTML]{9B9B9B} 0.56/0.91/0.94} &
  {\color[HTML]{9B9B9B} 0.45/0.89/0.95} &
  {\color[HTML]{9B9B9B} 0.43/0.76/0.89} &
  {\color[HTML]{9B9B9B} 0.03/0.07/0.45} &
  {\color[HTML]{9B9B9B} 4e4/-0.02/0.03} &
  {\color[HTML]{9B9B9B} 4e5/0.01/0.00} &
  {\color[HTML]{9B9B9B} 2.04/0.08/0.12} &
  {\color[HTML]{9B9B9B} 1e3/0.00/-0.02} \\
 &
   &
   &
  RNN &
  FC &
  {\color[HTML]{9B9B9B} 0.47/0.68/0.78} &
  {\color[HTML]{9B9B9B} 0.41/0.64/0.78} &
  {\color[HTML]{9B9B9B} 0.43/0.48/0.56} &
  {\color[HTML]{9B9B9B} 0.43/0.59/0.64} &
  {\color[HTML]{9B9B9B} 0.43/0.91/0.94} &
  {\color[HTML]{9B9B9B} 0.53/0.89/0.94} &
  {\color[HTML]{9B9B9B} 0.49/0.88/0.91} &
  {\color[HTML]{9B9B9B} 0.45/0.54/0.68} &
  {\color[HTML]{9B9B9B} 1e1/0.00/-0.03} &
  {\color[HTML]{9B9B9B} 5e4/0.00/-0.07} &
  {\color[HTML]{9B9B9B} 5e3/0.00/-0.06} &
  {\color[HTML]{9B9B9B} 2.15/-0.03/-0.06} \\
 &
  \multirow{-9}{*}{\rotatebox{90}{FT w/o PMI}} &
  \multirow{-3}{*}{MobileBERT} &
  GRU &
  FC &
  {\color[HTML]{9B9B9B} 0.48/0.69/0.78} &
  {\color[HTML]{9B9B9B} 0.43/0.64/0.74} &
  {\color[HTML]{9B9B9B} 0.00/0.49/0.49} &
  {\color[HTML]{9B9B9B} 0.00/0.22/0.23} &
  {\color[HTML]{9B9B9B} 0.52/0.89/0.93} &
  {\color[HTML]{9B9B9B} 0.55/0.89/0.94} &
  {\color[HTML]{9B9B9B} 0.40/0.87/0.92} &
  {\color[HTML]{9B9B9B} 0.02/0.03/.048} &
  {\color[HTML]{9B9B9B} 3e4/0.01/-0.05} &
  {\color[HTML]{9B9B9B} 6e5/-0.04/-0.03} &
  {\color[HTML]{9B9B9B} 6e2/-0.03/-0.10} &
  {\color[HTML]{9B9B9B} 5e5/0.03/0.01} \\ \cline{2-17}
 &
   &
   &
  LSTM &
  FC &
  {\color[HTML]{9B9B9B} 0.72/0.91/0.95} &
  {\color[HTML]{9B9B9B} 0.68/0.89/0.93} &
  {\color[HTML]{9B9B9B} 0.43/0.66/0.75} &
  {\color[HTML]{9B9B9B} 0.69/0.88/0.93} &
  {\color[HTML]{9B9B9B} 0.68/0.95/0.98} &
  {\color[HTML]{9B9B9B} 0.68/0.94/0.98} &
  {\color[HTML]{9B9B9B} 0.70/0.95/0.98} &
  {\color[HTML]{9B9B9B} 0.68/0.93/0.97} &
  {\color[HTML]{9B9B9B} 1.91/0.12/0.11} &
  {\color[HTML]{9B9B9B} 1.99/0.09/0.04} &
  {\color[HTML]{9B9B9B} 2.45/0.16/0.17} &
  {\color[HTML]{9B9B9B} 2.41/0.16/0.17} \\
 &
   &
   &
  RNN &
  FC &
  {\color[HTML]{9B9B9B} 0.72/0.90/0.94} &
  {\color[HTML]{9B9B9B} 0.69/0.89/0.92} &
  {\color[HTML]{9B9B9B} 0.66/0.88/0.94} &
  {\color[HTML]{9B9B9B} 0.67/0.88/0.92} &
  {\color[HTML]{9B9B9B} 0.66/0.95/0.98} &
  0.70{\color[HTML]{9B9B9B} /0.95/0.98} &
  {\color[HTML]{9B9B9B} 0.54/0.91/0.95} &
  {\color[HTML]{9B9B9B} 0.69/0.94/0.97} &
  {\color[HTML]{9B9B9B} 1.91/0.12/0.11} &
  {\color[HTML]{9B9B9B} 2.79/0.14/0.16} &
  {\color[HTML]{9B9B9B} 1.95/0.01/0.00} &
  {\color[HTML]{9B9B9B} 2.56/0.14/0.16} \\
 &
   &
  \multirow{-3}{*}{BERTBase} &
  GRU &
  FC &
  {\color[HTML]{9B9B9B} 0.72/0.89/0.95} &
  {\color[HTML]{9B9B9B} 0.70/0.90/0.95} &
  {0.70\color[HTML]{9B9B9B} /0.89/0.92} &
  {\color[HTML]{9B9B9B} 0.68/0.88/0.92} &
  {\color[HTML]{9B9B9B} 0.67/0.96/0.98} &
  {\color[HTML]{9B9B9B} 0.69/0.93/0.98} &
  {\color[HTML]{9B9B9B} 0.68/0.95/0.98} &
  {\color[HTML]{9B9B9B} 0.67/0.95/0.98} &
  {\color[HTML]{9B9B9B} 1.91/0.11/0.10} &
  {\color[HTML]{9B9B9B} 2.14/0.21/0.21} &
  {\color[HTML]{9B9B9B} 2.02/0.22/0.23} &
  {\color[HTML]{9B9B9B} 2.30/0.16/0.21} \\
 &
   &
   &
  LSTM &
  FC &
  {\color[HTML]{9B9B9B} 0.67/0.83/0.89} &
  {\color[HTML]{9B9B9B} 0.64/0.80/0.87} &
  {\color[HTML]{9B9B9B} 0.60/0.80/0.84} &
  {\color[HTML]{9B9B9B} 0.61/0.80/0.83} &
  {\color[HTML]{9B9B9B} 0.66/0.95/0.98} &
  {\color[HTML]{9B9B9B} 0.69/0.95/0.97} &
  {\color[HTML]{9B9B9B} 0.67/0.95/0.97} &
  {\color[HTML]{9B9B9B} 0.67/0.93/0.97} &
  {\color[HTML]{9B9B9B} 1.89/0.18/0.22} &
  {\color[HTML]{9B9B9B} 1.99/0.19/0.22} &
  {\color[HTML]{9B9B9B} 1.91/0.20/0.21} &
  {\color[HTML]{9B9B9B} 1.97/0.19/0.21} \\
 &
   &
   &
  RNN &
  FC &
  {\color[HTML]{9B9B9B} 0.68/0.85/0.90} &
  {\color[HTML]{9B9B9B} 0.63/0.80/0.86} &
  {\color[HTML]{9B9B9B} 0.59/0.80/0.86} &
  {\color[HTML]{9B9B9B} 0.59/0.79/0.84} &
  {\color[HTML]{9B9B9B} 0.69/0.95/0.97} &
  {\color[HTML]{9B9B9B} 0.68/0.95/0.97} &
  {\color[HTML]{9B9B9B} 0.69/0.95/0.97} &
  {\color[HTML]{9B9B9B} 0.66/0.95/0.97} &
  {\color[HTML]{9B9B9B} 1.89/0.17/0.19} &
  {\color[HTML]{9B9B9B} 2.03/0.16/0.19} &
  {\color[HTML]{9B9B9B} 1.94/0.18/0.20} &
  {\color[HTML]{9B9B9B} 1.96/0.19/0.23} \\
 &
   &
  \multirow{-3}{*}{TinyBERT} &
  GRU &
  FC &
  {\color[HTML]{9B9B9B} 0.67/0.83/0.90} &
  {\color[HTML]{9B9B9B} 0.62/0.79/0.85} &
  {\color[HTML]{9B9B9B} 0.61/0.81/0.86} &
  {\color[HTML]{9B9B9B} 0.56/0.78/0.86} &
  {\color[HTML]{9B9B9B} 0.70/}0.96{\color[HTML]{9B9B9B}/0.98} &
  {\color[HTML]{9B9B9B} 0.70/0.94/0.97} &
  {\color[HTML]{9B9B9B} 0.67/0.95/0.98} &
  {\color[HTML]{9B9B9B} 0.67/}0.95{\color[HTML]{9B9B9B}/0.97} &
  {\color[HTML]{9B9B9B} 1.96/0.14/0.18} &
  {\color[HTML]{9B9B9B} 1.95/0.20/0.23} &
  {\color[HTML]{9B9B9B} 1.96/0.16/0.18} &
  {\color[HTML]{9B9B9B} 1.99/0.19/0.23} \\
 &
   &
   &
  LSTM &
  FC &
  {\color[HTML]{9B9B9B} 0.72/0.89/0.94} &
  {\color[HTML]{9B9B9B} 0.68/0.88/0.93} &
  {\color[HTML]{9B9B9B} 0.66/0.88/0.92} &
  {\color[HTML]{9B9B9B} 0.64/0.86/0.89} &
  {\color[HTML]{9B9B9B} 0.66/0.95/0.98} &
  {\color[HTML]{9B9B9B} 0.68/0.95/0.97} &
  {\color[HTML]{9B9B9B} 0.68/0.95/0.98} &
  {\color[HTML]{9B9B9B} 0.68/0.93/0.97} &
  {\color[HTML]{9B9B9B} 1.93/0.13/0.14} &
  {\color[HTML]{9B9B9B} 2.32/0.19/0.22} &
  {\color[HTML]{9B9B9B} 2.24/0.15/0.18} &
  {\color[HTML]{9B9B9B} 2.12/0.15/0.18} \\
 &
   &
   &
  RNN &
  FC &
  {\color[HTML]{9B9B9B} 0.68/0.90/0.95} &
  {\color[HTML]{9B9B9B} 0.67/0.86/0.91} &
  {\color[HTML]{9B9B9B} 0.67/0.87/0.92} &
  {\color[HTML]{9B9B9B} 0.66/0.89/0.91} &
  {\color[HTML]{9B9B9B} 0.70/0.95/0.98} &
  {\color[HTML]{9B9B9B} 0.66/0.95/0.98} &
  {\color[HTML]{9B9B9B} 0.65/}0.96{\color[HTML]{9B9B9B}/0.97} &
  {\color[HTML]{9B9B9B} 0.67/0.93/0.96} &
  {\color[HTML]{9B9B9B} 1.93/0.15/0.18} &
  {\color[HTML]{9B9B9B} 2.11/0.18/0.23} &
  {\color[HTML]{9B9B9B} 2.03/0.14/0.14} &
  {\color[HTML]{9B9B9B} 2.19/0.15/0.18} \\
\multirow{-18}{*}{\rotatebox{90}{\textbf{Multimodal: 3-modal (our)}}} &
  \multirow{-9}{*}{\rotatebox{90}{\textbf{FT w/ PMI}}} &
  \multirow{-3}{*}{MobileBERT} &
  GRU &
  FC &
  {\color[HTML]{9B9B9B} 0.70/0.88/0.93} &
  {\color[HTML]{9B9B9B} 0.69/0.90/0.93} &
  {\color[HTML]{9B9B9B} 0.65/0.88/0.92} &
  {\color[HTML]{9B9B9B} 0.70/0.88/0.91} &
  {\color[HTML]{9B9B9B} 0.68/0.95/0.98} &
  {\color[HTML]{9B9B9B} 0.68/0.95/0.97} &
  {\color[HTML]{9B9B9B} 0.66/0.94/0.98} &
  {\color[HTML]{9B9B9B} 0.69/0.94/}0.98 &
  {\color[HTML]{9B9B9B} 1.96/0.13/0.17} &
  {\color[HTML]{9B9B9B} 2.27/0.16/0.20} &
  {\color[HTML]{9B9B9B} 2.36/0.13/0.15} &
  {\color[HTML]{9B9B9B} 2.28/0.18/0.23} \\ \hline
\end{tabular}

}
\end{table*}

\begin{figure}[t]
    \includegraphics[width=\linewidth]{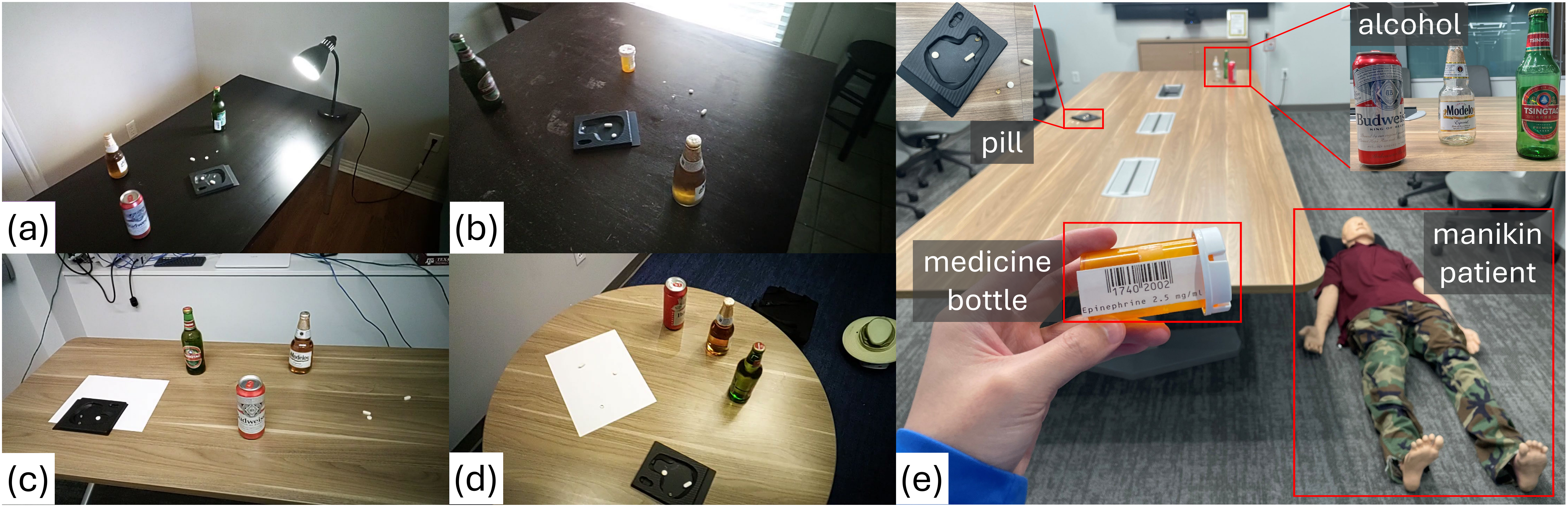}
    \caption{Image data collection room setup (a-d) and user study room setup (e).}
    \Description{Image data collection room setup (a-d) and user study room setup (e).}
    \label{fig:user_study_room}
\end{figure}

\section{Setup of image data collection room and user study room}
\label{sec:user_study_room}

As described in Section~\ref{sec:multimodal_processor_design}, Figure~\ref{fig:user_study_room}(a)-(e) are the room setups to collect the image dataset D4(image): 908 train set images are collected from (a)-(d), and 200 test set images from (e). Figure~\ref{fig:user_study_room}(e) is also the room setup used for the user study. 

\section{User study process and quantitative scores}
\label{sec:user_study}

\subsection{User study process}
\label{sec:user_study_process}

Before each user study, each EMT participant signed a consent form and received a brief orientation to the EMSGlass system and Google Glass hardware. The study began with a short hands-on tutorial introducing key hardware components (display prism, camera, microphone, and touchpad frame), and training participants to operate EMSGlass through simple tap and voice interactions. Participants were then guided through a demonstration scenario showing how EMSGlass transcribes symptoms, displays real-time vitals, detects scene objects (e.g., pills, alcohol bottles), and recommends corresponding EMS protocols, medicines, and dosages. Following the demonstration, participants performed a full simulation independently, interacting with the app to assess a manikin patient by describing symptoms, observing dynamic updates in vitals and recommendations, and completing the scenario by scanning medication labels. Each participant could repeat dry runs until comfortable with the workflow. This structured process ensured that all participants understood EMSGlass’s multimodal functionalities before proceeding to the formal evaluation.

\subsection{EMT user study assessment}
\label{sec:user_study_quant_assessment}

Figure~\ref{fig:user_study_quant_scores} shows quantitative assessment scores from 6 EMT users.





\end{document}